\pdfoutput=1

\documentclass[11pt]{article}

\usepackage{ACL2023}

\usepackage{times}
\usepackage{latexsym}

\usepackage[T1]{fontenc}

\usepackage[utf8]{inputenc}

\usepackage{microtype}

\usepackage{inconsolata}

\usepackage[utf8]{inputenc} 
\usepackage[T1]{fontenc}    
\usepackage{hyperref}       
\usepackage{url}            
\usepackage{booktabs}       
\usepackage{amsfonts}       
\usepackage{nicefrac}       
\usepackage{xcolor}         
\usepackage{xspace}
\usepackage{graphicx}
\usepackage{amsmath}    
\usepackage{bm}
\usepackage{arydshln}
\usepackage{subfigure}
\usepackage{enumitem}
\usepackage{multirow}
\usepackage{algorithm2e}
\usepackage{setspace}
\usepackage{amssymb}
\usepackage{caption}
\usepackage{cleveref}

\newcommand{\bmh}{{\bm h}}

\newcommand{\bmz}{{\bm z}}
\newcommand{\bmH}{{\bm H}}

\newcommand{\bmG}{{\bm G}}

\newcommand\mask{\texttt{[MASK]}\xspace}
\newcommand\cls{\texttt{[CLS]}\xspace}
\newcommand{\Ours}{\textsc{Patton}\xspace}

\newcommand*{\img}[1]{%
    \raisebox{-.02\baselineskip}{%
        \includegraphics[
        height=\baselineskip,
        width=\baselineskip,
        keepaspectratio,
        ]{#1}%
    }%
}

\title{\Ours \img{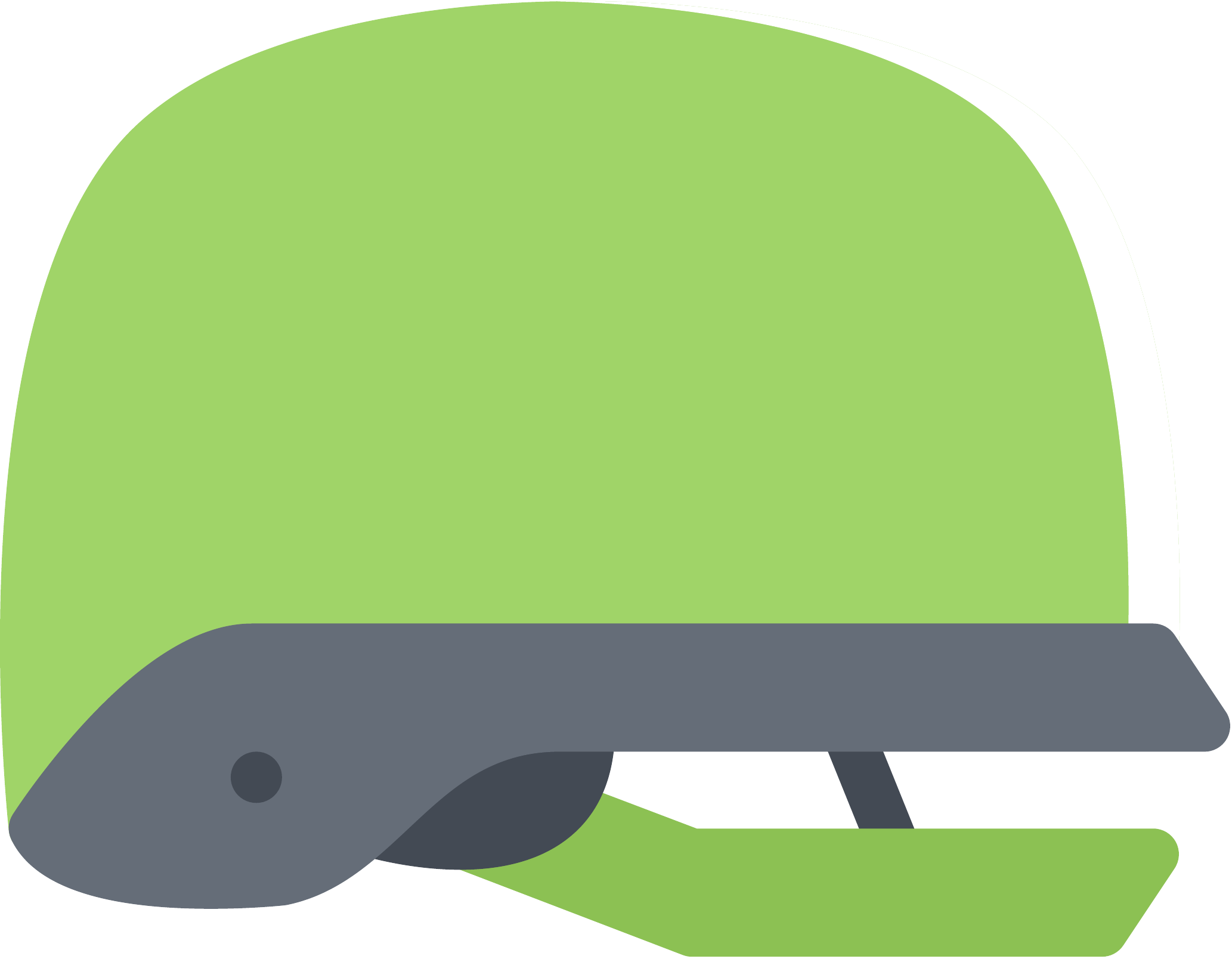}: Language Model Pretraining on Text-Rich Networks}

\author{Bowen Jin$^1$, Wentao Zhang$^1$, Yu Zhang$^1$, Yu Meng$^1$, \\ \textbf{Xinyang Zhang$^1$, Qi Zhu$^1$, Jiawei Han$^1$} \\
  $^1$University of Illinois at Urbana-Champaign, IL, USA \\
  \texttt{\small \{bowenj4,wentao4,yuz9,yumeng5,xz43,qiz3,hanj\}@illinois.edu}}

\begin{document}
\maketitle
\begin{abstract}
A real-world text corpus sometimes comprises not only text documents, but also semantic links between them (\textit{e.g.}, academic papers in a bibliographic network are linked by citations and co-authorships).
Text documents and semantic connections form a \textit{text-rich network}, which empowers a wide range of downstream tasks such as classification and retrieval.
However, pretraining methods for such structures are still lacking, making it difficult to build one generic model that can be adapted to various tasks on text-rich networks.
Current pretraining objectives, such as masked language modeling, purely model texts and do not take inter-document structure information into consideration.
To this end, we propose our \textit{\textbf{P}retr\textbf{A}ining on \textbf{T}ex\textbf{T}-Rich \textbf{N}etw\textbf{O}rk} framework \Ours.
\Ours \footnote{Code is available at \url{https://github.com/PeterGriffinJin/Patton}} includes two pretraining strategies: network-contextualized masked language modeling and masked node prediction, to capture the inherent dependency between textual attributes and network structure.
We conduct experiments on four downstream tasks in five datasets from both academic and e-commerce domains, where \Ours outperforms baselines significantly and consistently.

\end{abstract}

\section{Introduction}

Texts in the real world are often interconnected through links that can indicate their semantic relationships. For example, papers connected through citation links tend to be of similar topics; e-commerce items connected through co-viewed links usually have related functions. 
The texts and links together form a type of network called a \textit{text-rich network}, where documents are represented as nodes, and the edges reflect the links among documents.
Given a text-rich network, people are usually interested in various downstream tasks (\textit{e.g.,} document/node classification, document retrieval, and link prediction) \cite{zhang2019heterogeneous, wang2019heterogeneous, jinedgeformers}. For example, given a computer science academic network as context, it is intuitively appealing to automatically classify each paper \cite{kandimalla2021large}, find the authors of a new paper \cite{schulz2014exploiting}, and provide paper recommendations \cite{kuccuktuncc2012recommendation}. 
In such cases, pretraining a language model on a given text-rich network which can benefit a great number of downstream tasks inside this given network is highly demanded \cite{hu2020gpt}.

While there have been abundant studies on building generic pretrained language models \cite{peters2018deep,devlin2018bert,liu2019roberta, clark2020electra}, they are mostly designed for modeling texts exclusively, and do not consider inter-document structures.
Along another line of research, various network-based pretraining strategies are proposed in the graph learning domain to take into account structure information \cite{hu2019strategies, hu2020gpt}.
Yet, they focus on pretraining graph neural networks rather than language models and cannot easily model the rich textual semantic information in the networks.
To empower language model pretraining with network signals, LinkBERT \cite{yasunaga2022linkbert} is a pioneering study that puts two linked text segments together during pretraining so that they can serve as the context of each other. However, it simplifies the complex network structure into node pairs and does not model higher-order signals \cite{yang2021graphformers}.
Overall, both existing language model pretraining methods and graph pretraining methods fail to capture the rich contextualized textual semantic information hidden inside the complex network structure.

\begin{figure}[t]
\centering
\includegraphics[width=0.48\textwidth]{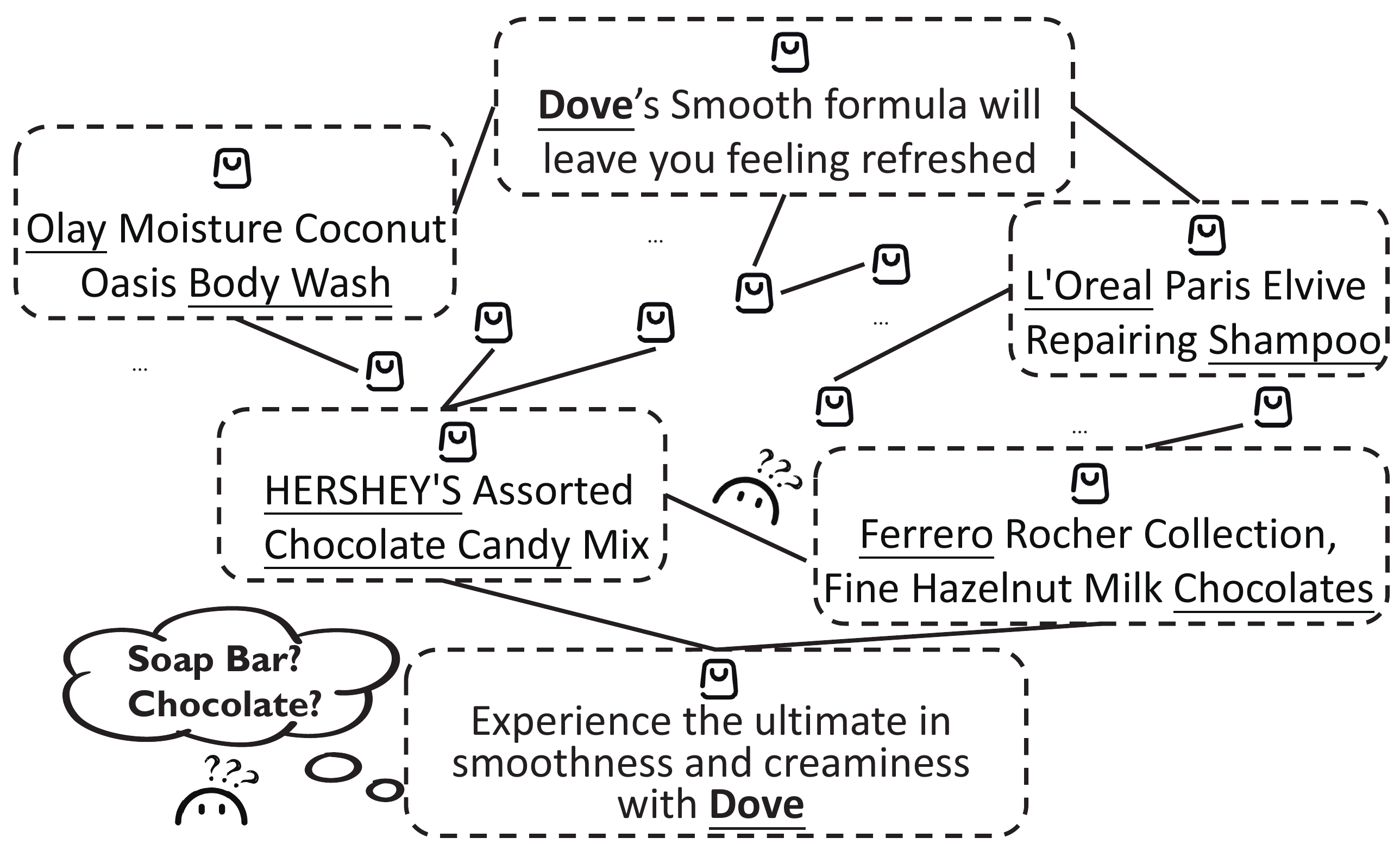}
\caption{An illustration of a text-rich network (a product item co-viewed network). 
At the \textit{token} level, from network neighbors, we can know that the ``Dove'' at the top is a personal care brand and the ``Dove'' at the bottom is a chocolate brand. At the \textit{document} level, referring to the edge in the middle, we can learn that the chocolate from ``Hershey's'' should have some similarity with the chocolate from ``Ferrero''.}\label{intro-figure}
\end{figure}

To effectively extract the contextualized semantics information, we propose to view the knowledge encoded inside the complex network structure from two perspectives: token-level and document-level.
At the \textit{token} level, neighboring documents can help facilitate the understanding of tokens. For example, in Figure \ref{intro-figure}, based on the text information of neighbors, we can know that the ``Dove'' at the top refers to a personal care brand, while the ``Dove'' at the bottom is a chocolate brand. At the \textit{document} level, the two connected nodes can have quite related overall textual semantics. For example, in Figure \ref{intro-figure}, the chocolate from ``Hershey's'' should have some similarity with the chocolate from ``Ferrero''. Absorbing such two-level hints in pretraining can help language models produce more effective representations which can be generalized to various downstream tasks.

To this end, we propose \Ours, a method to continuously pretrain language models on a given text-rich network. The key idea of \Ours is to leverage both textual information and network structure information to consolidate the pretrained language model's ability to understand tokens and documents.
Building on this idea, we propose two pretraining strategies:
1) Network-contextualized masked language modeling: We randomly mask several tokens within each node and train the language model to predict those masked tokens based on both in-node tokens and network neighbors' tokens. 2) Masked node prediction: We randomly mask some nodes inside the network and train the language model to correctly identify the masked nodes based on the neighbors' textual information.

We evaluate \Ours on both academic domain networks and e-commerce domain networks. To comprehensively understand how the proposed pretraining strategies can influence different downstream tasks, we conduct experiments on classification, retrieval, reranking, and link prediction. 

In summary, our contributions are as follows:
\begin{itemize}[leftmargin=*,nosep]
    \item We propose the problem of language model pretraining on text-rich networks.
    \item We design two strategies, network contextualized MLM and masked node prediction to train the language model to extract both token-level and document-level semantic correlation hidden inside the complex network structure.
    \item We conduct experiments on four downstream tasks in five datasets from different domains, where \Ours outperforms pure text/graph pretraining baselines significantly and consistently.
\end{itemize}

\section{Preliminaries}\label{sec::profdef}

\vspace{3px}
\noindent\textbf{Definition 2.1. Text-Rich Networks \cite{yang2021graphformers,jin2022heterformer}.}
A text-rich network can be denoted as $\mathcal{G}=(\mathcal{V}, \mathcal{E}, \mathcal{D})$, where $\mathcal{V}$, $\mathcal{E}$ and $\mathcal{D}$ are node set, edge set, and text set, respectively. Each $v_i\in \mathcal{V}$ is associated with some textual information $d_{v_i}\in \mathcal{D}$. For example, in an academic citation network, $v\in\mathcal{V}$ are papers, $e\in\mathcal{E}$ are citation edges, and $d\in\mathcal{D}$ are the content of the papers.
In this paper, we mainly focus on networks where the edges can provide semantic correlation between texts (nodes). For example, in a citation network, connected papers (cited papers) are likely to be semantically similar.

\vspace{3px}
\noindent\textbf{Problem Definition. (Language Model Pretraining on Text-rich Networks.)} 
Given a text-rich network $\mathcal{G}=(\mathcal{V}, \mathcal{E}, \mathcal{D})$, the task is to capture the self-supervised signal on $\mathcal{G}$ and obtain a $\mathcal{G}$-adapted language model $\mathcal{M_G}$.
The resulting language model $\mathcal{M_G}$ can be further finetuned on downstream tasks in $\mathcal{G}$, such as classification, retrieval, reranking, and link prediction, with only a few labels.

\section{\Ours}

\subsection{Model Architecture}\label{arch}

To jointly leverage text and network information in pretraining, we adopt the GNN-nested Transformer architecture (called GraphFormers) proposed in \cite{yang2021graphformers}.
In this architecture, GNN modules are inserted between Transformer layers. The forward pass of each GraphFormers layer is as follows.
\begin{equation}
    \bmz^{(l)}_x = \text{GNN}(\{\bmH^{(l)}_y\cls|y\in N_x\}),
\end{equation}
\begin{equation}
    \widetilde{\bmH}^{(l)}_x = \text{Concate}(\bmz^{(l)}_x, \bmH^{(l)}_x),
\end{equation}
\begin{equation}
    \begin{gathered}
        \widetilde{\bmH}^{(l)'}_x = \text{LN}(\bmH^{(l)}_x + \text{MHA}_{asy}(\widetilde{\bmH}^{(l)}_x)),
    \end{gathered}
\end{equation}
\begin{equation}
    \bmH^{(l+1)}_x = \text{LN}(\widetilde{\bmH}^{(l)'}_x + \text{MLP}(\widetilde{\bmH}^{(l)'}_x)),
\end{equation}
where $\bmH^{(l)}_x$ is token hidden states in the $l$-th layer for node $x$, $N_x$ is the network neighbor set of $x$, $\text{LN}$ is the layer normalization operation and $\text{MHA}_{asy}$ is the asymmetric multihead attention operation.
For more details, one can refer to \cite{yang2021graphformers}.

\begin{figure*}[t]
    \centering
    \includegraphics[width=0.95\linewidth]{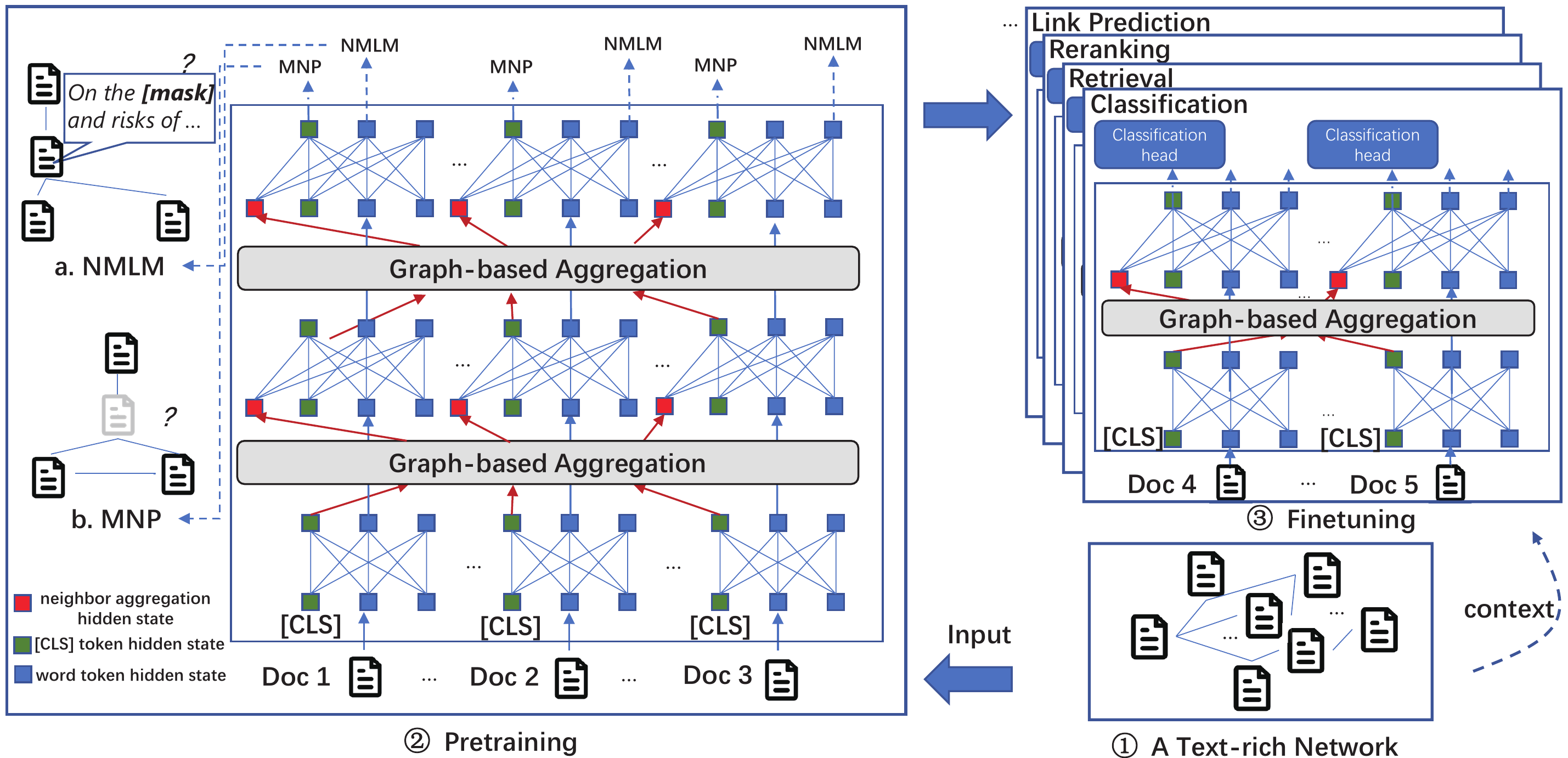}
    \caption{Overall pretraining and finetuning procedures for \Ours. We have two pretraining strategies: network-contextualized masked language modeling (NMLM) and masked node prediction (MNP). Apart from output layers, the same architectures are used in both pretraining and finetuning (in our experiment, we have 12 layers). The same pretrained model parameters are used to initialize models for different downstream tasks. During finetuning, all parameters are updated.}\label{fig:twhin-bert-framework}
\end{figure*}

\subsection{Pretraining \Ours}
We propose two strategies to help the language models understand text semantics on both the token level and the document level collaboratively from the network structure. The first strategy focuses on token-level semantics learning, namely network-contextualized masked language modeling; while the second strategy emphasizes document-level semantics learning, namely masked node prediction.

\paragraph{Strategy 1: Network-contextualized Masked Language Modeling (NMLM).}
Masked language modeling (MLM) is a commonly used strategy for language model pretraining \cite{devlin2018bert, liu2019roberta} and domain adaptation \cite{gururangan2020don}. It randomly masks several tokens in the text sequence and utilizes the surrounding unmasked tokens to predict them. 
The underlying assumption is that the semantics of each token can be reflected by its contexts.
Trained to conduct masked token prediction, the language model will learn to understand semantic correlation between tokens and capture the contextualized semantic signals. The mathematical formulation of MLM is as follows,
\begin{equation}
    \mathcal{L}_{\rm MLM} = -\sum_{i\in M_t} {\rm log}\ p(w_i|\bmH_i),
\end{equation}
where $M_t$ is a subset of tokens which are replaced by a special \mask token and $p(w_i|\bmH_i)$ is the output probability of a linear head $f_{\rm head}$ which gives predictions to $w_i$ (from the vocabulary $W$) based on contextualized token hidden states $\{\bmH_i\}$.

Such token correlation and contextualized semantics signals also exist and are even stronger in text-rich networks. Text from adjacent nodes in networks can provide auxiliary contexts for text semantics understanding. 
For example, given a paper talking about ``Transformers'' and its neighboring papers (cited papers) in the academic network on machine learning, we can infer that ``Transformers'' here is a deep learning model rather than an electrical engineering component by reading the text within both the given paper and the neighboring papers. 
In order to fully capture the textual semantic signals in the network, the language model needs to not only understand the in-node text token correlation but also be aware of the cross-node semantic correlation. 

We extend the original in-node MLM to network-contextualized MLM, so as to facilitate the language model to understand both in-node token correlation and network-contextualized text semantic relatedness. The training objective is shown as follows.
\begin{equation}
    \begin{gathered}
    \mathcal{L}_{\rm NMLM} = -\sum_{i\in M_t} {\rm log}\ p(w_i|\bmH_x, \bmz_x), \\
    p(w_i|\bmH_x, \bmz_x) = \text{softmax}(\bm{q}^\top_{w_i}\bmh_i),
    \end{gathered}
\end{equation}
where $\bmz_x$ denotes the network contextualized token hidden state in Section \ref{arch} and $\bmh_i=\bmH^{(L)}_x[i]$ (if $i$ is inside node $x$). $L$ is the number of layers. $\bm{q}_{w_i}$ refers to the MLM prediction head for $w_i$. Since the calculation of $\bmh_i$ is based on $\bmH_x$ and $\bmz_x$, the likelihood will be conditioned on $\bmH_x$ and $\bmz_x$.

\paragraph{Strategy 2: Masked Node Prediction (MNP).}
While network-contextualized MLM focuses more on token-level semantics understanding, we propose a new strategy called ``masked node prediction'', which helps the language model understand the underlying document-level semantics correlation hidden in the network structure.

Concretely, we dynamically hold out a subset of nodes from the network ($M_v\subseteq V$), mask them, and train the language model to predict the masked nodes based on the adjacent network structure. 
\begin{equation}
    \begin{gathered}
    \mathcal{L}_{\rm MNP} = -\sum_{v_j\in M_v} {\rm log}\ p(v_j|\bmG_{v_j}),\\
    p(v_j|\bmG_{v_j}) = \text{softmax}(\bmh^\top_{v_j}\bmh_{N_{v_j}})
    \end{gathered}
\end{equation}
where $\bmG_{v_j}=\{\bmh_{v_k}|v_k\in N_{v_j}\}$ are the hidden states of the neighbor nodes in the network and $N_{v_j}$ is the set of neighbors of $v_j$. In particular, we treat the hidden state of the last layer of \cls as a representation of node level, that is, $\bmh_{v_j}=\bmH^{(L)}_{v_j}\cls$.

By performing the task, the language model will absorb document semantic hints hidden inside the network structure (\textit{e.g.}, contents between cited papers in the academic network can be quite semantically related, and text between co-viewed items in the e-commerce network can be highly associated).

However, directly optimizing masked node prediction can be computationally expensive since we need to calculate the representations for all neighboring nodes and candidate nodes for one prediction.
To ease the computation overload, we prove that the masked node prediction task can be theoretically transferred to a computationally cheaper pairwise link prediction task.


\noindent\textbf{Theorem 3.2.1.} \textit{Masked node prediction is equivalent to pairwise link prediction.}

\noindent\textit{Proof:} Given a set of masked nodes $M_v$, the likelihood of predicting the masked nodes is
\begin{equation*}
    \begin{split}
    & \prod_{v_\mask \in M_v} p(v_\mask = v_i | v_k \in N_{v_\mask}) \\
    & \propto \prod_{v_\mask \in M_v} p(v_k \in N_{v_\mask}|v_\mask = v_i) \\
    & = \prod_{v_\mask \in M_v} \prod_{v_k \in N_{v_\mask}} p(v_k|v_\mask = v_i) \\
    & = \prod_{v_\mask \in M_v} \prod_{v_k \in N_{v_\mask}} p(v_k \longleftrightarrow v_i) \\
    \end{split}
\end{equation*}
In the above proof, the first step relies on the Bayes' rule, and we have the assumption that all nodes appear uniformly in the network, \textit{i.e.,} $p(v_\mask = v_i)=p(v_\mask = v_j)$. In the second step, we have the conditional independence assumption of neighboring nodes generated given the center node, \textit{i.e.}, $p(v_k, v_s|v_\mask = v_i)=p(v_k|v_\mask = v_i)\cdot p(v_s|v_\mask = v_i)$.

As a result, the masked node prediction objective can be simplified into a pairwise link prediction objective, which is

\begin{small}
    \begin{equation}
        \begin{split}
        & \mathcal{L}_{\rm MNP} = -\sum_{v_j\in M_v}\sum_{v_k\in N_{v_j}} {\rm log}\ p(v_j\leftrightarrow v_k) \\
        &= -\sum_{v_j\in{M_v}}\sum_{v_k\in {N}_{v_j}} \text{log}\frac{\text{exp}(\bmh^\top_{v_j} \bmh_{v_k})}{\text{exp}(\bmh^\top_{v_j} \bmh_{v_k})+\sum_{u'}\text{exp}(\bmh^\top_{v_j} \bmh_{v_u'})},
        \end{split}
    \end{equation}
\end{small}
where $v_u'$ stands for a random negative sample. In our implementation, we use ``in-batch negative samples'' \cite{karpukhin2020dense} to reduce the encoding cost.

\paragraph{Joint Pretraining.}
To pretrain \Ours, we optimize the NMLM objective and the MNP objective jointly:
\begin{equation}
    \mathcal{L}=\mathcal{L}_{\rm NMLM} + \mathcal{L}_{\rm MNP}.
\end{equation}
This joint objective will unify the effects of NMLM and MNP, which encourages the model to conduct network-contextualized token-level understanding and network-enhanced document-level understanding, facilitating the joint modeling of texts and network structures. We will show in Section \ref{ablation} that the joint objective achieves superior performance in comparison with using either objective alone.

\subsection{Finetuning \Ours}
Last, we describe how to finetune \Ours for downstream tasks involving encoding for text in the network and text not in the network. 
For text in the network (thus with neighbor information), we will feed both the node text sequence and the neighbor text sequences into the model; 
while for texts not in the network (thus neighbor information is not available), we will feed the text sequence into the model and leave the neighbor text sequences blank. 
For both cases, the final layer hidden state of \cls is used as text representation following \cite{devlin2018bert} and \cite{liu2019roberta}.

\section{Experiments}

\subsection{Experimental Settings}

\paragraph{Dataset.}We perform experiments on both academic networks from Microsoft Academic Graph (MAG) \cite{sinha2015overview} and e-commerce networks from Amazon \cite{mcauley2015image}. In academic networks, nodes are papers and there will be an edge between two papers if one cites the other; while in e-commerce networks, nodes correspond to items, and item nodes are linked if they are frequently co-viewed by users. 
Since MAG and Amazon both have multiple domains, we select three domains from MAG and two domains from Amazon. In total, five datasets are used in the evaluation (\textit{i.e.}, MAG-Mathematics, MAG-Geology, MAG-Economics, Amazon-Clothes and Amazon-Sports). The statistics of all the datasets can be found in Table \ref{tab:dataset}.
Fine-classes are all the categories in the network-associated node category taxonomy (MAG taxonomy and Amazon product catalog), while coarse-classes are the categories at the first layer of the taxonomy.

\begin{table}[t]
    \centering
    \caption{Dataset Statistics.}\label{tab:dataset}
    \resizebox{0.45\textwidth}{!}{
    \begin{tabular}{c|cccc}
    \hline
         Dataset & \#Nodes  & \#Edges & \#Fine-Classes & \#Coarse-Classes\\
    \hline
    Mathematics & 490,551 & 2,150,584 & 14,271  & 18\\
    Geology & 431,834  & 1,753,762 & 7,883 & 17\\
    Economics & 178,670  & 1,042,253 & 5,205 & 40\\
    Clothes & 889,225  & 7,876,427 & 2,771 & 9\\
    Sports & 314,448 & 3,461,379 & 3,034  & 16\\
    \bottomrule
    \end{tabular}}
\end{table}

\paragraph{Pretraining Setup.}

The model is trained for 5/10/30 epochs (depending on the size of the network) on 4 Nvidia A6000 GPUs with a total batch size of 512. 
We set the peak learning rate as 1e-5. 
NMLM pretraining uses the standard 15\% \mask ratio. 
For our model and all baselines, we adopt a 12-layer architecture. 
More details can be found in the Appendix \ref{App:pretrain_setup}.

\paragraph{Baselines.}
We mainly compare our method with two kinds of baselines, off-the-shelf pretrained language models and language model continuous pretraining methods. 
The first category includes BERT \cite{devlin2018bert}, SciBERT \cite{Beltagy2019SciBERT}, SPECTER \cite{specter2020cohan}, SimCSE \cite{gao2021simcse}, LinkBERT \cite{yasunaga2022linkbert} and vanilla GraphFormers \cite{yang2021graphformers}.
BERT \cite{devlin2018bert} is a language model pretrained with masked language modeling and next sentence prediction objectives on Wikipedia and BookCorpus. SciBERT \cite{Beltagy2019SciBERT} utilizes the same pretraining strategies as BERT but is trained on 1.14 million paper abstracts and full text from Semantic Scholar. SPECTER \cite{specter2020cohan} is a language model continuously pretrained from SciBERT with a contrastive objective on 146K scientific papers. SimCSE \cite{gao2021simcse} is a contrastive learning framework and we perform the experiment with the models pretrained from both unsupervised settings (Wikipedia) and supervised settings (NLI). LinkBERT \cite{yasunaga2022linkbert} is a language model pretrained with masked language modeling and document relation prediction objectives on Wikipedia and BookCorpus. GraphFormers \cite{yang2021graphformers} is a GNN-nested Transformer and we initialize it with the BERT checkpoint for a fair comparison.
The second category includes several continuous pretraining methods \cite{gururangan2020don, gao2021simcse}. We perform continuous masked language modeling starting from the BERT checkpoint (denoted as BERT.MLM) and the SciBERT checkpoint (denoted as SciBERT.MLM) on our data, respectively. We also perform in-domain supervised contrastive pretraining with the method proposed in \cite{gao2021simcse} (denoted as SimCSE.in-domain).

\paragraph{Ablation Setup.}
For academic networks, we pretrain our model starting from the BERT-base \footnote{https://huggingface.co/bert-base-uncased} checkpoint (\Ours) and the SciBERT \footnote{https://huggingface.co/allenai/scibert\_scivocab\_uncased} checkpoint (Sci\Ours) respectively; while for e-commerce networks, we pretrain our model from BERT-base only (\Ours). Furthermore, we conduct ablation studies to validate the effectiveness of both the NMLM and the MNP strategies. 
The pretrained model with NMLM removed and that with MNP removed are called ``w/o NMLM'' and ``w/o MNP'', respectively. In academic networks, the ablation study is done on Sci\Ours, while in e-commerce networks, it is done on \Ours.

We demonstrate the effectiveness of our framework on four downstream tasks, including classification, retrieval, reranking, and link prediction.

\subsection{Classification}

\begin{table*}[t]
\centering
\caption{Experiment results on Classification. We show the $\text{mean}_\text{std}$ of three runs for all the methods.}
\setlength{\tabcolsep}{2.5mm}
\scalebox{0.62}{
\begin{tabular}{lcccccccccc}
\toprule
\multicolumn{1}{c}{\multirow{2}{*}{Method}} & \multicolumn{2}{c}{\textbf{Mathematics}} & \multicolumn{2}{c}{\textbf{Geology}} & \multicolumn{2}{c}{\textbf{Economics}} & \multicolumn{2}{c}{\textbf{Clothes}} & \multicolumn{2}{c}{\textbf{Sports}} \\
\multicolumn{1}{c}{}  & \textbf{Macro-F1} & \textbf{Micro-F1} & \textbf{Macro-F1} & \textbf{Micro-F1} & \textbf{Macro-F1} & \textbf{Micro-F1} & \textbf{Macro-F1} & \textbf{Micro-F1} & \textbf{Macro-F1} & \textbf{Micro-F1} \\ \midrule
BERT   & $18.14_{0.07}$ &	$22.04_{0.32}$ & $21.97_{0.87}$	& $29.63_{0.36}$ & $14.17_{0.08}$ &	$19.77_{0.12}$ & $45.10_{1.47}$ &	$68.54_{2.25}$ & $31.88_{0.23}$ &	$34.58_{0.56}$       \\
GraphFormers  &  $18.69_{0.52}$ &	$23.24_{0.46}$ & $22.64_{0.92}$	& $31.02_{1.16}$ & $13.68_{1.03}$	& $19.00_{1.44}$     & $46.27_{1.92}$ &	$68.97_{2.46}$ & $43.77_{0.63}$ &	$50.47_{0.78}$             \\
SciBERT  & $23.50_{0.64}$ &	$23.10_{2.23}$ & $29.49_{1.25}$ &	$37.82_{1.89}$ & $15.91_{0.48}$	& $21.32_{0.66}$    & - & - & - & -          \\
SPECTER   &  $23.37_{0.07}$ &	$29.83_{0.96}$ & $30.40_{0.48}$	& $38.54_{0.77}$ & $16.16_{0.17}$ &	$19.84_{0.47}$      & - & - & - & -            \\
SimCSE (unsup)  &  $20.12_{0.08}$ &	$26.11_{0.39}$ & $38.78_{0.19}$	& $38.55_{0.17}$ & $14.54_{0.26}$ &	$19.07_{0.43}$      & $42.70_{2.32}$	& $58.72_{0.34}$   & $41.91_{0.85}$ &	$59.19_{0.55}$ \\
SimCSE (sup)  & $20.39_{0.07}$ &	$25.56_{0.00}$ & $25.66_{0.28}$	 & $33.89_{0.40}$ & $15.03_{0.53}$ &	$18.64_{1.32}$      & $52.82_{0.87}$ &	$75.54_{0.98}$    & $46.69_{0.10}$ & $59.19_{0.55}$         \\
LinkBERT  & $15.78_{0.91}$ &	$19.75_{1.19}$ & $24.08_{0.58}$	 & $31.32_{0.04}$ & $12.71_{0.12}$ &	$16.39_{0.22}$      & $44.94_{2.52}$ &	$65.33_{4.34}$    & $35.60_{0.33}$ & $38.30_{0.09}$         \\
\midrule
BERT.MLM  &  $23.44_{0.39}$ & $31.75_{0.58}$ & $36.31_{0.36}$ & $48.04_{0.69}$ & $16.60_{0.21}$ & $22.71_{1.16}$ & $46.98_{0.84}$ & $68.00_{0.84}$ & $62.21_{0.13}$ & $75.43_{0.74}$                   \\
SciBERT.MLM  & $23.34_{0.42}$ & $30.11_{0.97}$ & $36.94_{0.28}$ & $46.54_{0.40}$ & $16.28_{0.38}$ & $21.41_{0.81}$ & - & - & - & -\\
SimCSE.in-domain   &  $25.15_{0.09}$ & $29.85_{0.20}$ & $38.91_{0.08}$ & $48.93_{0.1
4}$ & $18.08_{0.22}$ & $23.79_{0.44}$ & $57.03_{0.20}$ & $80.16_{0.31}$ & $65.57_{0.35}$ & $75.22_{0.18}$                   \\
\midrule
\Ours  &  $\textbf{27.58}_{0.03}$ & $\textbf{32.82}_{0.01}$ & ${39.35}_{0.06}$ & ${48.19}_{0.15}$ & ${19.32}_{0.05}$ & ${25.12}_{0.05}$  & $\textbf{60.14}_{0.28}$ & $\textbf{84.88}_{0.09}$ & $\textbf{67.57}_{0.08}$ & $\textbf{78.60}_{0.15}$                 \\
Sci\Ours &  ${27.35}_{0.04}$ & ${31.70}_{0.01}$ & $\textbf{39.65}_{0.10}$ & $\textbf{48.93}_{0.06}$ & $\textbf{19.91}_{0.08}$ & $\textbf{25.68}_{0.32}$  & \textbf{-} & \textbf{-} & \textbf{-} & \textbf{-}                 \\
\hdashline
w/o NMLM & $25.91_{0.45}$  & $27.79_{2.07}$ &  $38.78_{0.19}$	 & $48.48_{0.17}$ & $18.86_{0.23}$ & $24.25_{0.26}$ & $56.68_{0.24}$ & $80.27_{0.17}$ & $65.83_{0.28}$ & $76.24_{0.54}$ \\
w/o MNP & $24.79_{0.65}$  & $29.44_{1.50}$ & $38.00_{0.73}$ & $47.82_{1.06}$ & $18.69_{0.59}$ & $25.63_{1.44}$ & $47.35_{1.20}$ & $68.50_{2.60}$ & $64.23_{1.53}$ & $76.03_{1.67}$ \\
\bottomrule
\end{tabular}
}
\label{tab:classification}
\end{table*}

In this section, we conduct experiments on 8-shot coarse-grained classification for nodes in the networks. We use the final layer hidden state of \cls token from language models as the representation of the node and feed it into a linear layer classifier to obtain the prediction result. Both the language model and the classifier are finetuned. The experimental results are shown in Table \ref{tab:classification}. From the result, we can find that: 1) \Ours and Sci\Ours consistently outperform baseline methods; 2) Continuous pretraining method (BERT.MLM, SciBERT.MLM, SimCSE.in-domain, \Ours, and Sci\Ours) can have better performance than off-the-shelf PLMs, which demonstrates that domain shift exists between the pretrained PLM domain and the target domain, and the adaptive pretraining on the target domain is necessary.
More detailed information on the task can be found in Appendix \ref{apx:class}.

\subsection{Retrieval}

The retrieval task corresponds to 16-shot fine-grained category retrieval, where given a node, we want to retrieve category names for it from a very large label space. We follow the widely-used DPR \cite{karpukhin2020dense} pipeline to finetune all the models. In particular, the final layer hidden states of \cls token are utilized as dense representations for both node and label names. Negative samples retrieved from BM25 are used as hard negatives. The results are shown in Table \ref{tab:retrieval}. From the result, we can have the following observations: 1) \Ours and Sci\Ours consistently outperform all the baseline methods; 2) Continuously pretrained models can be better than off-the-shelf PLMs in many cases (SciBERT and SPECTER perform well on Mathematics and Economics since their pretrained corpus includes a large number of Computer Science papers, which are semantically close to Mathematics and Economics papers) and can largely outperform traditional BM25.
More detailed information on the task can be found in Appendix \ref{apx:retrieval}.

\begin{table*}[ht]
\centering
\caption{Experiment results on Retrieval. We show the $\text{mean}_\text{std}$ of three runs for all the methods.}
\setlength{\tabcolsep}{2.5mm}
\scalebox{0.62}{
\begin{tabular}{lcccccccccc}
\toprule
\multicolumn{1}{c}{\multirow{2}{*}{Method}} & \multicolumn{2}{c}{\textbf{Mathematics}} & \multicolumn{2}{c}{\textbf{Geology}} & \multicolumn{2}{c}{\textbf{Economics}} & \multicolumn{2}{c}{\textbf{Clothes}} & \multicolumn{2}{c}{\textbf{Sports}} \\
\multicolumn{1}{c}{}  & \textbf{R@50} & \textbf{R@100} & \textbf{R@50} & \textbf{R@100} & \textbf{R@50} & \textbf{R@100} & \textbf{R@50} & \textbf{R@100} & \textbf{R@50} & \textbf{R@100} \\ 
\midrule
BM25   & $20.76$ &  $24.55$	 & $19.02$	& $20.92$ & $19.14$ & $22.49$ & $15.76$ & $15.88$   & $22.00$ & $23.96$      \\
\midrule
BERT   & $16.73_{0.17}$ &	$22.66_{0.18}$ & $18.82_{0.39}$	& $25.94_{0.39}$ & $23.95_{0.25}$ & $31.54_{0.21}$ & $40.77_{1.68}$ & $50.40_{1.41}$   & $32.37_{1.09}$ & $43.32_{0.96}$     \\
GraphFormers  &  $16.65_{0.12}$ &	$22.41_{0.10}$ & $18,92_{0.60}$	& $25.94_{0.39}$ & $24.48_{0.36}$	& $32.16_{0.40}$     & $41.77_{2.05}$ & $51.26_{2.27}$   & $32.39_{0.89}$ & $43.29_{1.12}$            \\
SciBERT  & $24.70_{0.17}$ &	$33.55_{0.31}$ & $23.71_{0.89}$ &	$30.94_{0.95}$ & $29.80_{0.66}$	& $38.66_{0.52}$   & - & - & - & -          \\
SPECTER   &  $23.86_{0.25}$ &	$31.11_{0.31}$ & $26.56_{1.05}$	& $34.04_{1.32}$ & $31.26_{0.15}$ &	$40.79_{0.11}$      & - & - & - & -            \\
SimCSE (unsup)  &  $17.91_{0.26}$ & $23.19_{0.29}$ & $20.45_{0.20}$ & $26.82_{0.26}$ & $25.83_{0.23}$ & $33.42_{0.28}$     & $44.90_{0.35}$ & $54.76_{0.38}$    & $38.81_{0.35}$ & $49.30_{0.44}$              \\
SimCSE (sup)  & $20.29_{0.41}$ &	$26.23_{0.51}$ & $22.34_{0.49}$	 & $29.63_{0.55}$ & $28.07_{0.38}$ &  $36.51_{0.37}$  & $44.69_{0.59}$ & $54.70_{0.77}$ &   $40.31_{0.43}$ & $50.55_{0.41}$          \\
LinkBERT  & $17.25_{0.30}$ &	$23.21_{0.47}$ & $17.14_{0.75}$	 & $23.05_{0.74}$ & $22.69_{0.30}$ &	$30.77_{0.36}$      & $28.66_{2.97}$ &	$37.79_{3.82}$    & $31.97_{0.54}$ & $41.77_{0.67}$         \\
\midrule
BERT.MLM  &  $20.69_{0.21}$ & $27.17_{0.25}$ & $32.13_{0.36}$ & $41.74_{0.42}$ & $27.13_{0.04}$ & $36.00_{0.14}$ & $52.41_{1.71}$ & $63.72_{1.79}$ & $54.10_{0.81}$ & $63.14_{0.83}$                   \\
SciBERT.MLM  &  $20.65_{0.21}$ & $27.67_{0.32}$ & $31.65_{0.71}$ & $40.52_{0.76}$ & $29.23_{0.67}$ & $39.18_{0.73}$ & - & - & - & - \\
SimCSE.in-domain   &  $24.54_{0.05}$ & $31.66_{0.09}$ & $33.97_{0.07}$ & $44.09_{0.19}$ & $28.44_{0.31}$ & $37.81_{0.27}$ & $61.42_{0.84}$ & $72.25_{0.86}$ & $53.77_{0.22}$ &   $63.73_{0.30}$               \\
\midrule
\Ours  &  \textbf{$27.44_{0.15}$} & \textbf{$34.97_{0.21}$} & \textbf{$34.94_{0.23}$} & \textbf{$45.01_{0.28}$} & \textbf{$32.10_{0.51}$} & \textbf{$42.19_{0.62}$}  & \textbf{$\textbf{68.62}_{0.38}$} & \textbf{$\textbf{77.54}_{0.19}$} & \textbf{$\textbf{58.63}_{0.31}$} & \textbf{$\textbf{68.53}_{0.55}$}                 \\
Sci\Ours &  $\textbf{31.40}_{0.52}$ & \textbf{$\textbf{40.38}_{0.66}$} & \textbf{$\textbf{40.69}_{0.52}$} & \textbf{$\textbf{51.31}_{0.48}$} & \textbf{$\textbf{35.82}_{0.69}$} & \textbf{${46.05}_{0.69}$}  & \textbf{-} & \textbf{-} & \textbf{-} & \textbf{-}                 \\
\hdashline
w/o NMLM &  $30.85_{0.14}$ & $39.89_{0.23}$ & $39.29_{0.07}$ & $49.59_{0.11}$ & $35.17_{0.31}$ & $\textbf{46.07}_{0.20}$ & $65.60_{0.26}$ & $75.19_{0.32}$ & $57.05_{0.14}$ & $67.22_{0.12}$ \\
w/o MNP &  $22.47_{0.07}$ & $30.20_{0.15}$ & $31.28_{0.89}$ & $40.54_{0.97}$ & $29.54_{0.36}$ & $39.57_{0.57}$ & $60.20_{0.73}$ & $69.85_{0.52}$ & $51.73_{0.41}$ & $60.35_{0.78}$ \\
\bottomrule
\end{tabular}
}
\label{tab:retrieval}
\end{table*}

\subsection{Reranking}
The reranking task corresponds to the 32-shot fine-grained category reranking. We first adopt BM25 \cite{robertson2009probabilistic} and exact matching as the retriever to obtain a candidate category name list for each node. Then, the models are asked to rerank all the categories in the list based on their similarity to the given node text.
The way to encode the node and category names is the same as that in retrieval. Unlike retrieval, reranking tests the ability of the language model to distinguish among candidate categories at a fine-grained level.
The results are shown in Table \ref{tab:reranking}.
From the result, we can find that \Ours and Sci\Ours consistently outperform all baseline methods, demonstrating that our pretraining strategies allow the language model to better understand fine-grained semantic similarity. 
More detailed information on the task can be found in Appendix \ref{apx:reranking}.

\begin{table*}[ht]
\centering
\caption{Experiment results on Reranking. We show the $\text{mean}_\text{std}$ of three runs for all the methods.}
\setlength{\tabcolsep}{2.5mm}
\scalebox{0.6}{
\begin{tabular}{lcccccccccc}
\toprule
\multicolumn{1}{c}{\multirow{2}{*}{Method}} & \multicolumn{2}{c}{\textbf{Mathematics}} & \multicolumn{2}{c}{\textbf{Geology}} & \multicolumn{2}{c}{\textbf{Economics}} & \multicolumn{2}{c}{\textbf{Clothes}} & \multicolumn{2}{c}{\textbf{Sports}} \\
\multicolumn{1}{c}{}  & \textbf{NDCG@5} & \textbf{NDCG@10} & \textbf{NDCG@5} & \textbf{NDCG@10} & \textbf{NDCG@5} & \textbf{NDCG@10} & \textbf{NDCG@5} & \textbf{NDCG@10} & \textbf{NDCG@5} & \textbf{NDCG@10} \\ \midrule
BERT  &	$37.15_{0.64}$ & $44.76_{0.59}$	& $56.59_{1.18}$ & $68.21_{0.96}$ & 	$42.65_{0.70}$ & $53.55_{0.76}$
 &  $62.19_{0.63}$  & $72.00_{0.70}$ &  $44.68_{0.56}$  &  $57.54_{0.55}$   \\
GraphFormers  &  $37.85_{0.32}$ & $47.89_{0.69}$ &  $58.32_{1.22}$ & $69.91_{1.19}$ &  $41.82_{0.65}$ & $52.67_{0.76}$ & $62.11_{0.87}$  & $72.02_{0.73}$ &  $44.49_{0.71}$   &   $57.35_{0.50}$       \\
SciBERT  & $40.73_{0.50}$ & $53.22_{0.51}$ & 	$57.04_{1.05}$ & $69.47_{0.92}$ & $43.24_{0.79}$ & $55.22_{0.67}$  & - & - & - & -          \\
SPECTER    &	$38.95_{0.67}$ & $52.17_{0.71}$ &  $57.79_{0.69}$ & $69.57_{0.46}$ & 	$43.41_{1.10}$  &  $55.80_{1.02}$  & - & - & - & -            \\
SimCSE (unsup)  &   $32.34_{0.43}$ & $42.59_{0.44}$
 &  $49.60_{1.04}$ & $61.51_{1.03}$ &  $36.37_{0.67}$  &  $47.18_{0.76}$  &  $57.03_{1.27}$  & $68.16_{1.04}$ &  $43.29_{0.16}$    &    $55.41_{0.09}$     \\
SimCSE (sup)  &	$34.85_{0.60}$ & $44.76_{0.59}$ &  $48.07_{0.54}$ & $59.79_{0.51}$ &   $37.01_{0.40}$  & $48.05_{0.44}$ &  $52.74_{0.55}$  & $64.28_{0.52}$ &  $42.00_{0.09}$    &    $53.92_{0.13}$    \\
LinkBERT   &	$38.50_{1.15}$ &  $50.74_{1.12}$  &  $59.57_{0.96}$ & $71.41_{0.93}$  & 	$44.00_{1.12}$   &  $55.78_{0.95}$ & 	$58.24_{1.93}$  & $70.48_{1.58}$  &   $48.45_{1.02}$   &  $61.63_{1.01}$    \\
\midrule
BERT.MLM  &   $39.24_{0.47}$ &  $51.18_{0.35}$ &  $60.58_{0.29}$ & $72.52_{0.28}$ &  $44.30_{0.68}$ & $55.84_{0.69}$
  &  $60.51_{0.31}$ & $71.36_{0.28}$  &  $45.70_{4.49}$   &   $57.08_{4.60}$             \\
SciBERT.MLM & $39.03_{0.48}$ & $52.34_{0.39}$ &  $62.01_{0.55}$ &  $74.58_{0.47}$ &  $46.43_{0.21}$ & $58.60_{0.21}$ & - & - & - & -                   \\
SimCSE.in-domain    & $40.37_{0.30}$ & $53.80_{0.24}$
  & $61.13_{0.75}$ &  $73.89_{0.57}$ & $45.27_{0.13}$ & $58.33_{0.13}$
  &  $64.81_{0.49}$ & $75.77_{0.24}$  &  $50.05_{0.62}$  &   $62.56_{0.29}$              \\
\midrule
\Ours   & ${42.08}_{0.17}$ & ${55.30}_{0.17}$  &  ${61.41}_{0.62}$ & $74.02_{0.49}$ &  ${46.52}_{0.53}$  & $59.25_{0.44}$
  & $\textbf{66.26}_{0.81}$ & 
  $\textbf{77.01}_{0.55}$ &  $\textbf{52.16}_{0.44}$     &   $64.96_{0.37}$        \\
Sci\Ours & $\textbf{47.10}_{0.49}$ & $\textbf{60.86}_{0.55}$
  &  $\textbf{63.48}_{0.25}$ & $\textbf{75.86}_{0.18}$
  & $\textbf{51.19}_{0.33}$  & $\textbf{63.86}_{0.34}$ & \textbf{-} & \textbf{-} & \textbf{-} & \textbf{-}                 \\
\hdashline
w/o NMLM  & $41.43_{0.16}$ & $55.28_{0.21}$ &  $62.84_{1.79}$ & $75.36_{1.43}$ &  $46.05_{2.04}$ & $59.39_{1.91}$  &  $63.71_{1.11}$ & $74.75_{0.81}$
  &  $52.12_{0.13}$  &  $\textbf{65.35}_{0.14}$ \\
w/o MNP   &  $43.56_{0.53}$ & $57.14_{0.52}$  &  $62.42_{0.47}$ &  $74.91_{0.40}$ &   $48.07_{0.30}$ &  $60.57_{0.32}$ &   $63.88_{0.47}$ & $74.01_{0.36}$
 &  $47.81_{0.56}$  & $59.68_{0.54}$ \\
\bottomrule
\end{tabular}
}
\label{tab:reranking}
\end{table*}

\subsection{Link Prediction}
In this section, we perform the 32-shot link prediction for nodes in the network. Language models are asked to give a prediction on whether there should exist an edge between two nodes. It is worth noting that the edge semantics here (``author overlap'' \footnote{Two papers have at least one same author.} for academic networks and ``co-purchased'' for e-commerce networks) are different from those in pretraining (``citation'' for academic networks and ``co-viewed'' for e-commerce networks). We utilize the final layer \cls token hidden state as node representation and conduct in-batch evaluations.
The results are shown in Table \ref{tab:lp}.
From the result, we can find that \Ours and Sci\Ours can outperform baselines and ablations in most cases, which shows that our pretraining strategies can help the language model extract knowledge from the pretrained text-rich network and apply it to the new link type prediction. 
More detailed information on the task can be found in Appendix \ref{apx:lp}.

\begin{table*}[ht]
\centering
\caption{Experiment results on Link Prediction. We show the $\text{mean}_\text{std}$ of three runs for all the methods.}
\setlength{\tabcolsep}{2.5mm}
\scalebox{0.6}{
\begin{tabular}{lcccccccccc}
\toprule
\multicolumn{1}{c}{\multirow{2}{*}{Method}} & \multicolumn{2}{c}{\textbf{Mathematics}} & \multicolumn{2}{c}{\textbf{Geology}} & \multicolumn{2}{c}{\textbf{Economics}} & \multicolumn{2}{c}{\textbf{Clothes}} & \multicolumn{2}{c}{\textbf{Sports}} \\
\multicolumn{1}{c}{}  & \textbf{PREC@1} & \textbf{MRR} & \textbf{PREC@1} & \textbf{MRR} & \textbf{PREC@1} & \textbf{MRR} & \textbf{PREC@1} & \textbf{MRR} & \textbf{PREC@1} & \textbf{MRR} \\ \midrule
BERT   & $6.60_{0.16}$ &	$12.96_{0.34}$ & $6.24_{0.76}$	& $12.96_{1.34}$ & $4.12_{0.08}$ &	$9.23_{0.15}$ & $24.17_{0.41}$ & $34.20_{0.45}$    & $16.48_{0.45}$ & $25.35_{0.52}$       \\
GraphFormers  &  $6.91_{0.29}$ &	$13.42_{0.34}$ & $6.52_{1.17}$	& $13.34_{1.81}$ & $4.16_{0.21}$	& $9.28_{0.28}$   & $23.79_{0.69}$ & $33.79_{0.66}$    & $16.69_{0.36}$ & $25.74_{0.48}$             \\
SciBERT  & $14.08_{0.11}$ &	$23.62_{0.10}$ & $7.15_{0.26}$ &	$14.11_{0.39}$ & $5.01_{1.04}$	& $10.48_{1.79}$    & - & - & - & -          \\
SPECTER   &  $13.44_{0.5}$ &	$21.73_{0.65}$ & $6.85_{0.22}$	& $13.37_{0.34}$ & $6.33_{0.29}$ &	$12.41_{0.33}$      & - & - & - & -            \\
SimCSE (unsup)  &  $9.85_{0.10}$ & $16.28_{0.12}$ & $7.47_{0.55}$ & $14.24_{0.89}$ & $5.72_{0.26}$ & $11.02_{0.34}$      & $30.51_{0.09}$ & $40.40_{0.10}$    & $22.99_{0.07}$ & $32.47_{0.06}$              \\
SimCSE (sup)  & $10.35_{0.52}$ &	$17.01_{0.72}$ & $10.10_{0.04}$	 & $17.80_{0.07}$ & $5.72_{0.26}$ &  $11.02_{0.34}$  & $35.42_{0.06}$ & $46.07_{0.06}$    & $27.07_{0.15}$ & $37.44_{0.16}$           \\
LinkBERT  & $8.05_{0.14}$ &	$13.91_{0.09}$ & $6.40_{0.14}$	 & $12.99_{0.17}$ & $2.97_{0.08}$ &	$6.79_{0.15}$      & $30.33_{0.56}$ &	$39.59_{0.64}$    & $19.83_{0.09}$ & $28.32_{0.04}$         \\
\midrule
BERT.MLM  &  $17.55_{0.25}$ & $29.22_{0.26}$ & $14.13_{0.19}$ & $25.36_{0.20}$ & $9.02_{0.09}$ & $16.72_{0.15}$ & $42.71_{0.31}$ & $54.54_{0.35}$ & $29.36_{0.09}$ & $41.60_{0.05}$  \\
SciBERT.MLM  &  $22.44_{0.08}$ & $34.22_{0.05}$ & $16.22_{0.03}$ & $27.02_{0.07}$ & $9.80_{0.00}$ & $17.72_{0.01}$ & - & - & - & -                   \\
SimCSE.in-domain   &  $33.55_{0.05}$ & $46.07_{0.07}$ & $24.56_{0.06}$ & $36.89_{0.11}$ & $16.77_{0.10}$ & $26.93_{0.01}$ & $\textbf{60.41}_{0.03}$ & $\textbf{71.86}_{0.06}$ & $49.17_{0.04}$ & $63.48_{0.03}$                   \\
\midrule
\Ours  &  ${70.41}_{0.11}$ & ${80.21}_{0.04}$ & ${44.76}_{0.05}$ & ${57.71}_{0.04}$ & ${57.04}_{0.05}$ & ${68.35}_{0.04}$  & $58.59_{0.12}$ & $70.12_{0.12}$ & $46.68_{0.09}$ & $60.96_{0.23}$                 \\
Sci\Ours &  $\textbf{71.22}_{0.17}$ & $\textbf{80.79}_{0.10}$ & $\textbf{44.95}_{0.24}$ & $\textbf{57.84}_{0.25}$ & $\textbf{57.36}_{0.26}$ & $\textbf{68.71}_{0.31}$  & \textbf{-} & \textbf{-} & \textbf{-} & \textbf{-}                 \\
\hdashline
w/o NMLM & $71.04_{0.13}$  & $80.60_{0.07}$ & $44.33_{0.23}$ & $57.29_{0.22}$ & $56.64_{0.25}$ & $68.12_{0.16}$ & $60.30_{0.03}$ & $71.67_{0.07}$ & $\textbf{49.72}_{0.06}$ & $\textbf{63.76}_{0.04}$ \\
w/o MNP &  $63.06_{0.23}$ & $74.26_{0.11}$ & $33.84_{0.60}$ & $47.02_{0.65}$ & $44.46_{0.03}$ & $57.05_{0.04}$ & $49.62_{0.06}$ & $61.61_{0.01}$ & $36.05_{0.20}$ & $49.78_{0.25}$ \\
\bottomrule
\end{tabular}
}
\label{tab:lp}
\end{table*}

\subsection{Ablation Study}\label{ablation}
We perform ablation studies to validate the effectiveness of the two strategies in Tables \ref{tab:classification}-\ref{tab:lp}.
The full method is better than each ablation version in most cases, except R@100 on Economy retrieval, NDCG@10 on Sports reranking, and link prediction on Amazon datasets, which indicates the importance of both strategies.

\subsection{Pretraining Step Study}

\begin{figure*}[t]
    \centering
    \subfigure[Classification]{\includegraphics[width=0.24\textwidth]{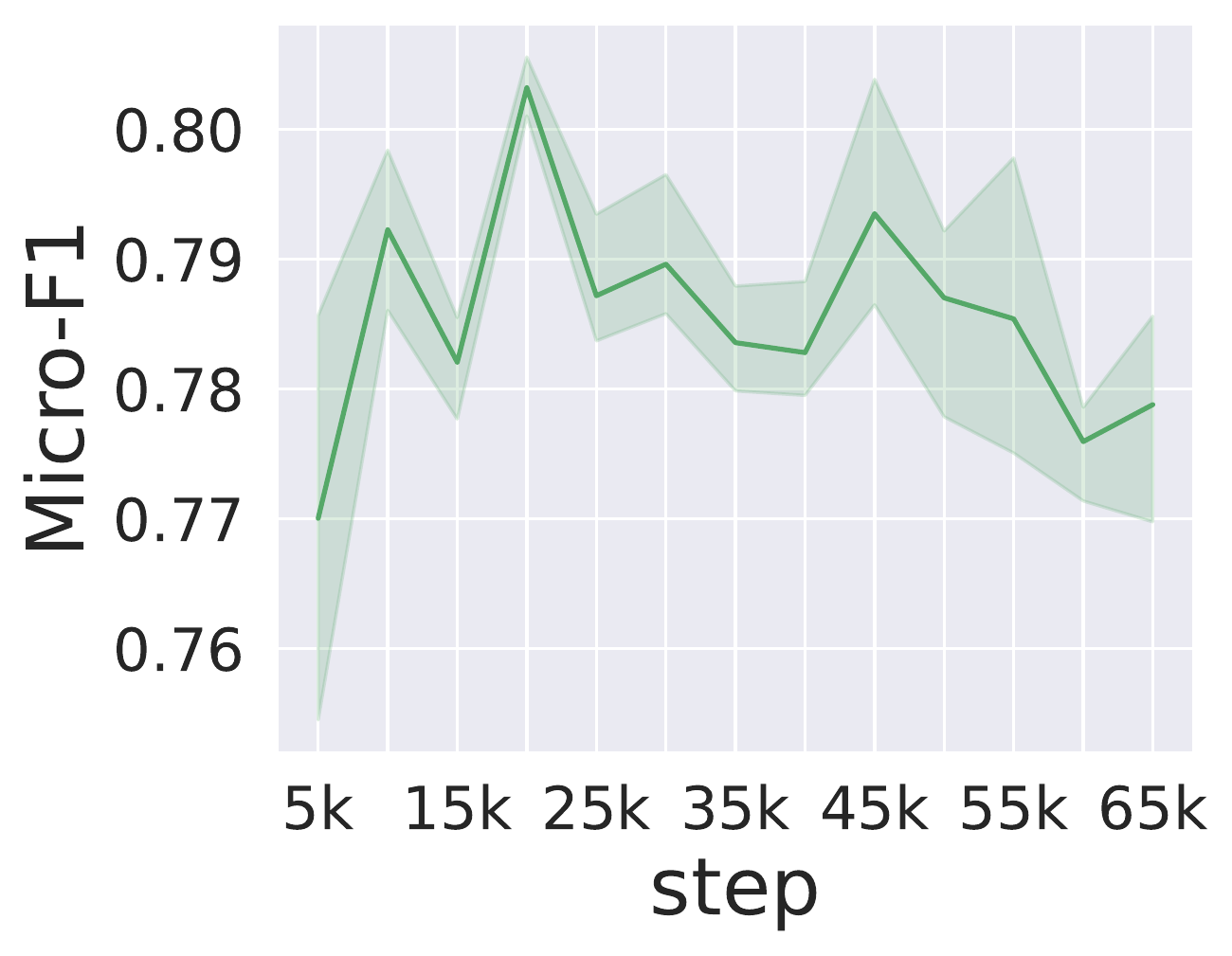}}
    \subfigure[Retrieval]{\includegraphics[width=0.24\textwidth]{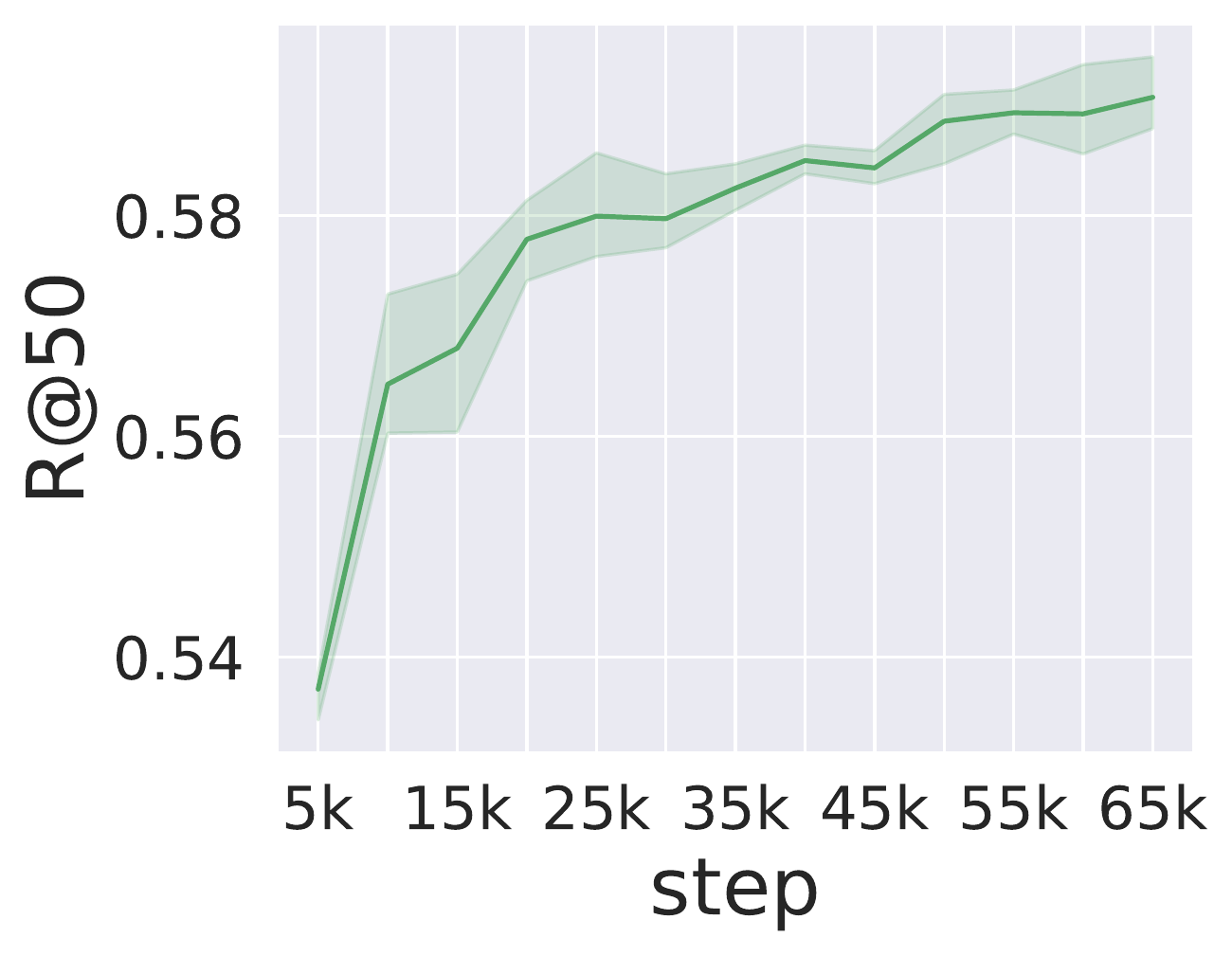}} 
    \subfigure[Reranking]{\includegraphics[width=0.24\textwidth]{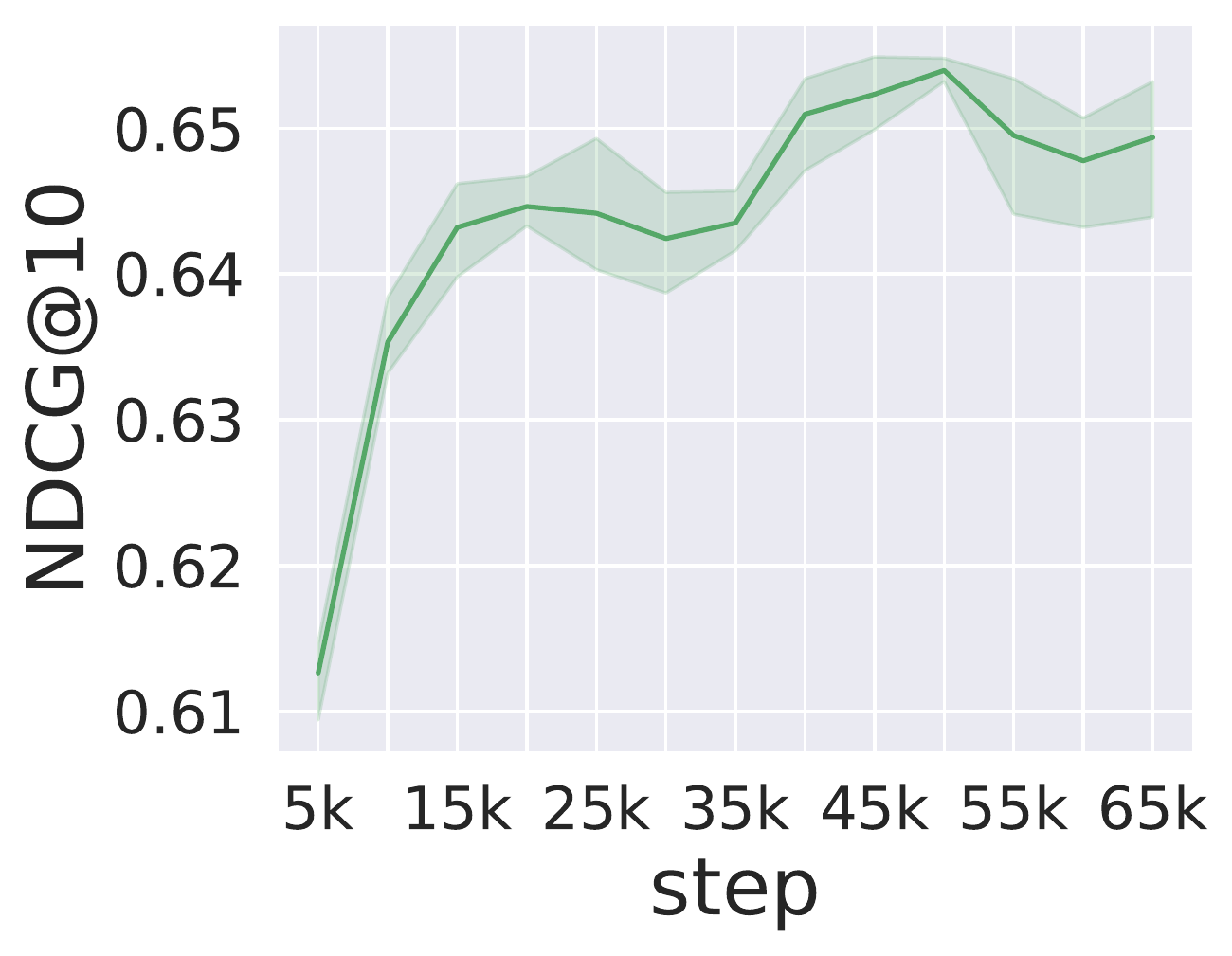}}
    \subfigure[Link Prediction]{\includegraphics[width=0.24\textwidth]{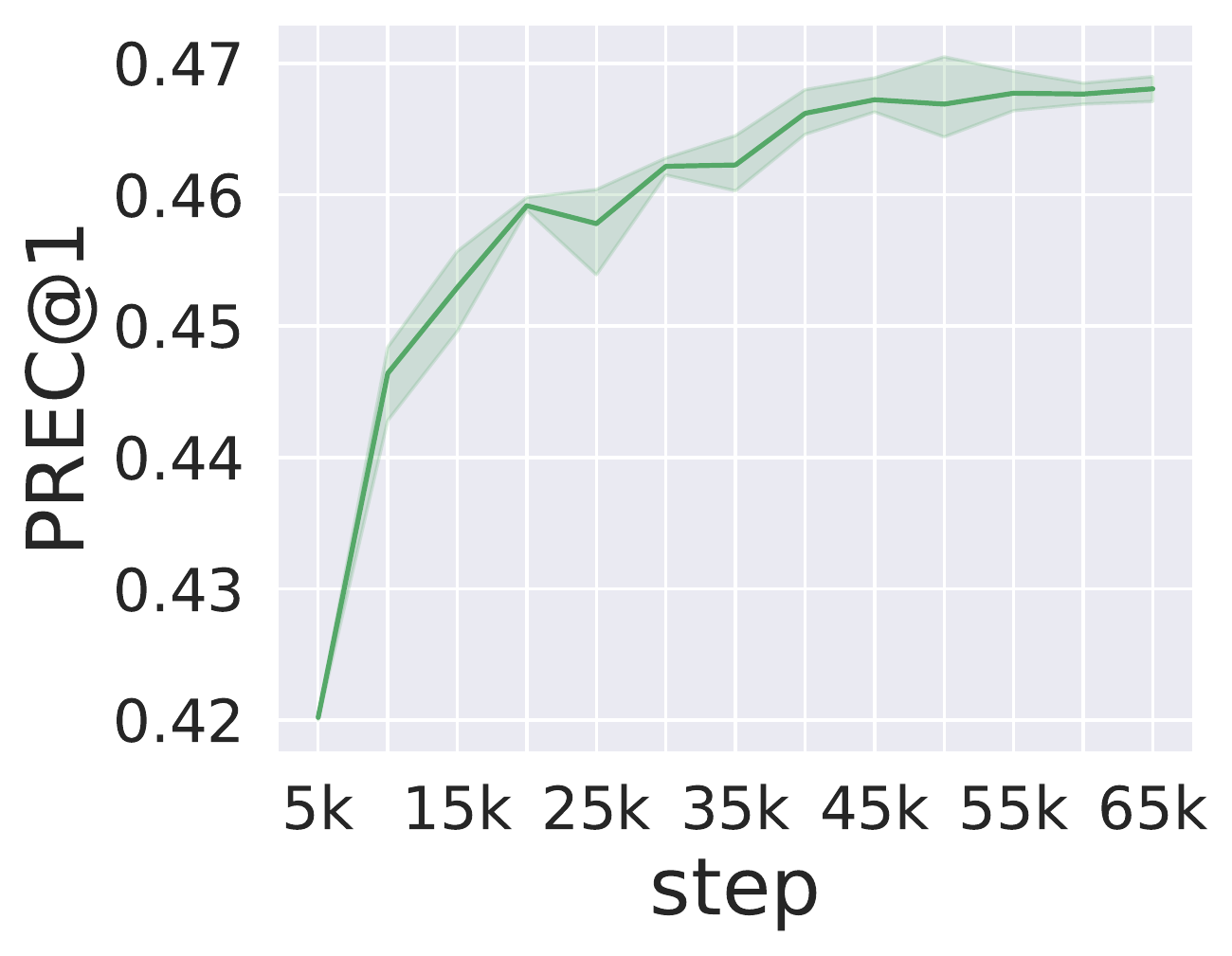}}
    \caption{Pretrain step study on Amazon-Sports. The downstream performance on retrieval, reranking and link prediction generally improves when pretrained for longer, while the performance on classification improves and then drops.}\label{fig:pretrain-step}
\end{figure*}

We conduct an experiment on the Sports dataset to study how the pretrained checkpoint at different pretraining steps can perform on downstream tasks. The result is shown in Figure \ref{fig:pretrain-step}.
From the figure, we can find that: 1) The downstream performance on retrieval, reranking, and link prediction generally improves as the pretraining step increases. This means that the pretrained language model can learn more knowledge, which can benefit these downstream tasks from the pretraining text-rich network as the pretraining step increases. 2) The downstream performance on classification increases and then decreases. The reason is that for downstream classification, when pretrained for too long, the pretrained language model may overfit the given text-rich network, which will hurt classification performance.

\subsection{Scalability Study}

We run an experiment on Sports to study the time complexity and memory complexity of the proposed pretraining strategies. The model is pretrained for 10 epochs on four Nvidia A6000 GPU devices with a total training batch size set as 512. We show the result in Table \ref{tab:scale}. From the result, we can find that: 1) Pretraining with the MNP strategy is faster and memory cheaper than pretraining with the NMLM strategy. 2) Combining the two strategies together will not increase the time complexity and memory complexity too much, compared with NMLM pretraining only.


Further model studies on finetune data size can be found in Appendix \ref{apx:finetune}.

\begin{table}[t]
    \centering
    \caption{Time scalability and memory scalability study on Amazon-Sports.}\label{tab:scale}
    \resizebox{0.45\textwidth}{!}{
    \begin{tabular}{lccc}
    \hline
         Attribute & NMLM & MNP & NMLM+MNP\\
    \hline
    Time & 15h 37min & 14h 53min & 15h 39min \\
    Memory & 32,363MB & 30,313MB & 32,365MB  \\
    \bottomrule
    \end{tabular}}
\end{table}

\section{Attention Map Study}\label{apx:att-map}

\begin{figure*}[t]
    \centering
    \subfigure[Classification w/o pretrain]{\includegraphics[width=0.24\textwidth]{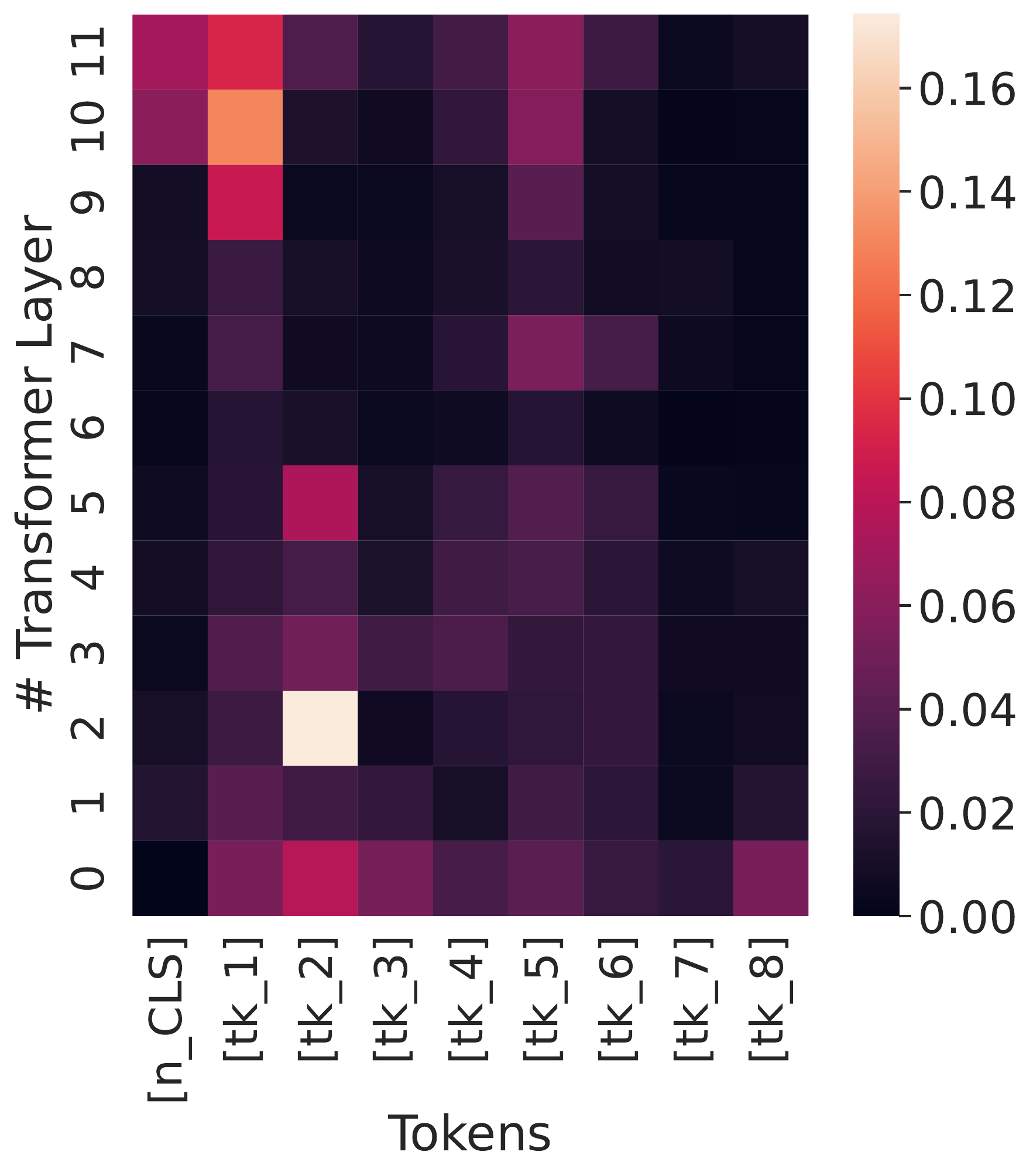}}
    \subfigure[Retrieval w/o pretrain]{\includegraphics[width=0.24\textwidth]{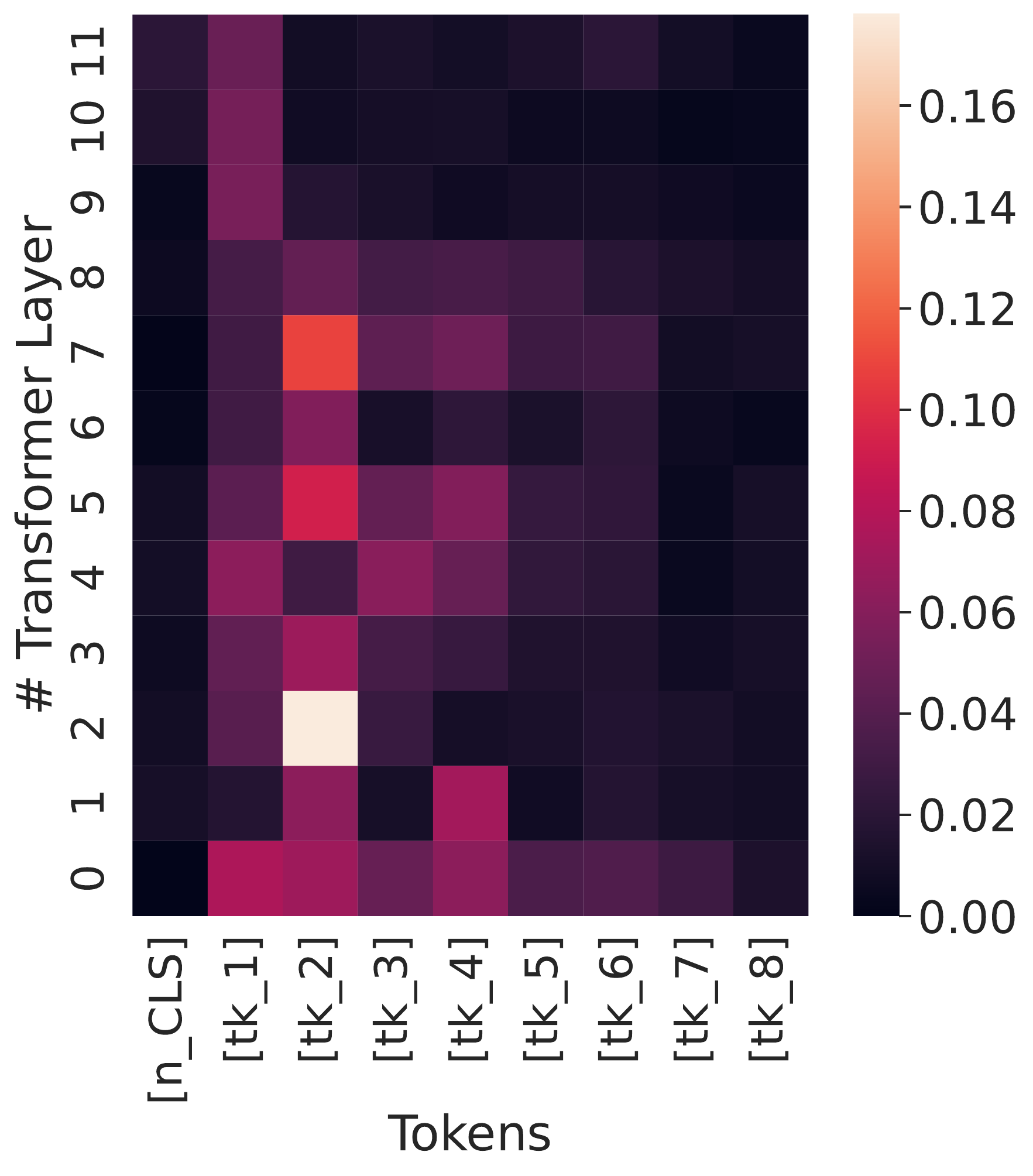}} 
    \subfigure[Reranking w/o pretrain]{\includegraphics[width=0.24\textwidth]{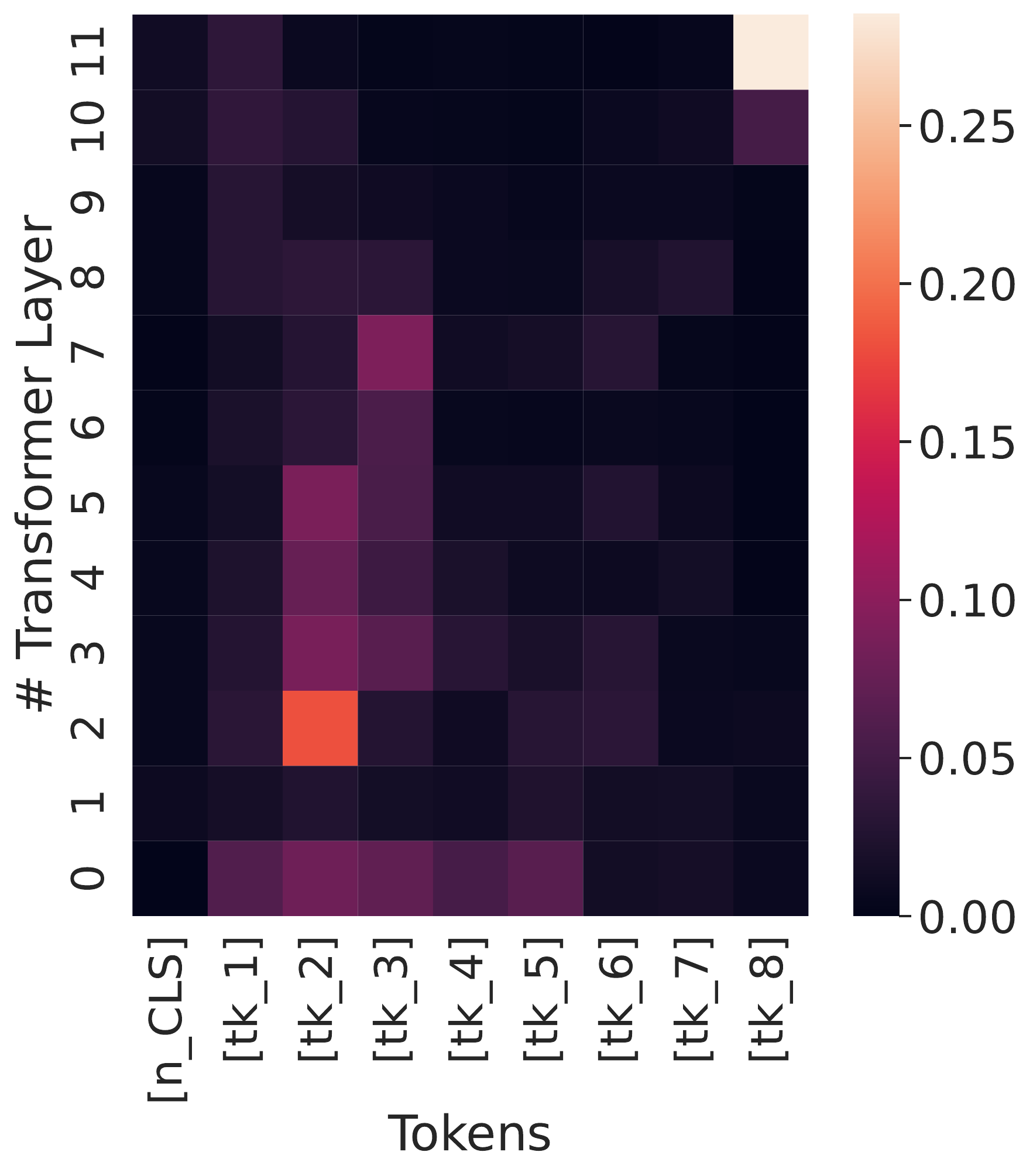}}
    \subfigure[Link Prediction w/o pretrain]{\includegraphics[width=0.24\textwidth]{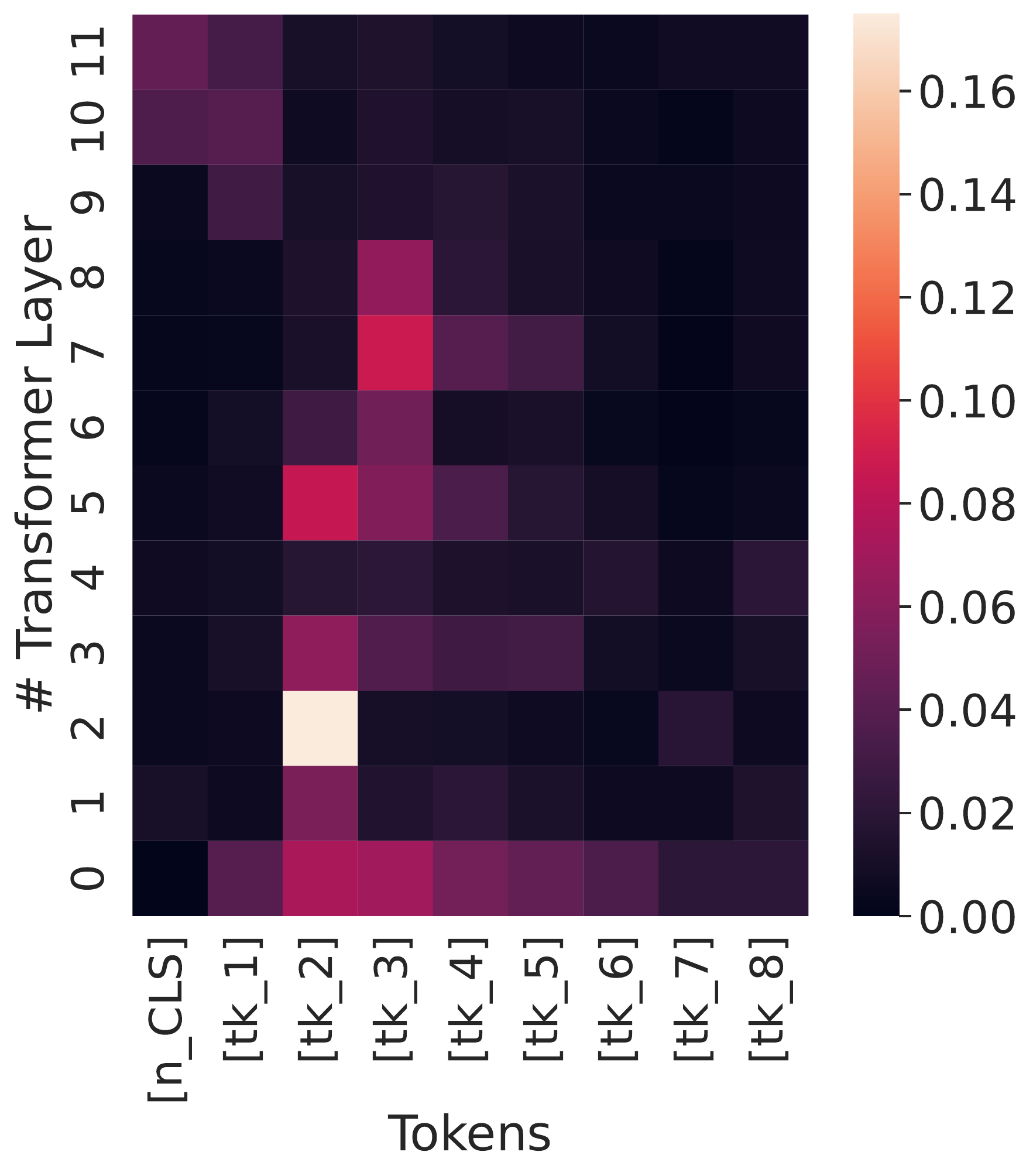}}
    \subfigure[Classification w/ pretrain]{\includegraphics[width=0.24\textwidth]{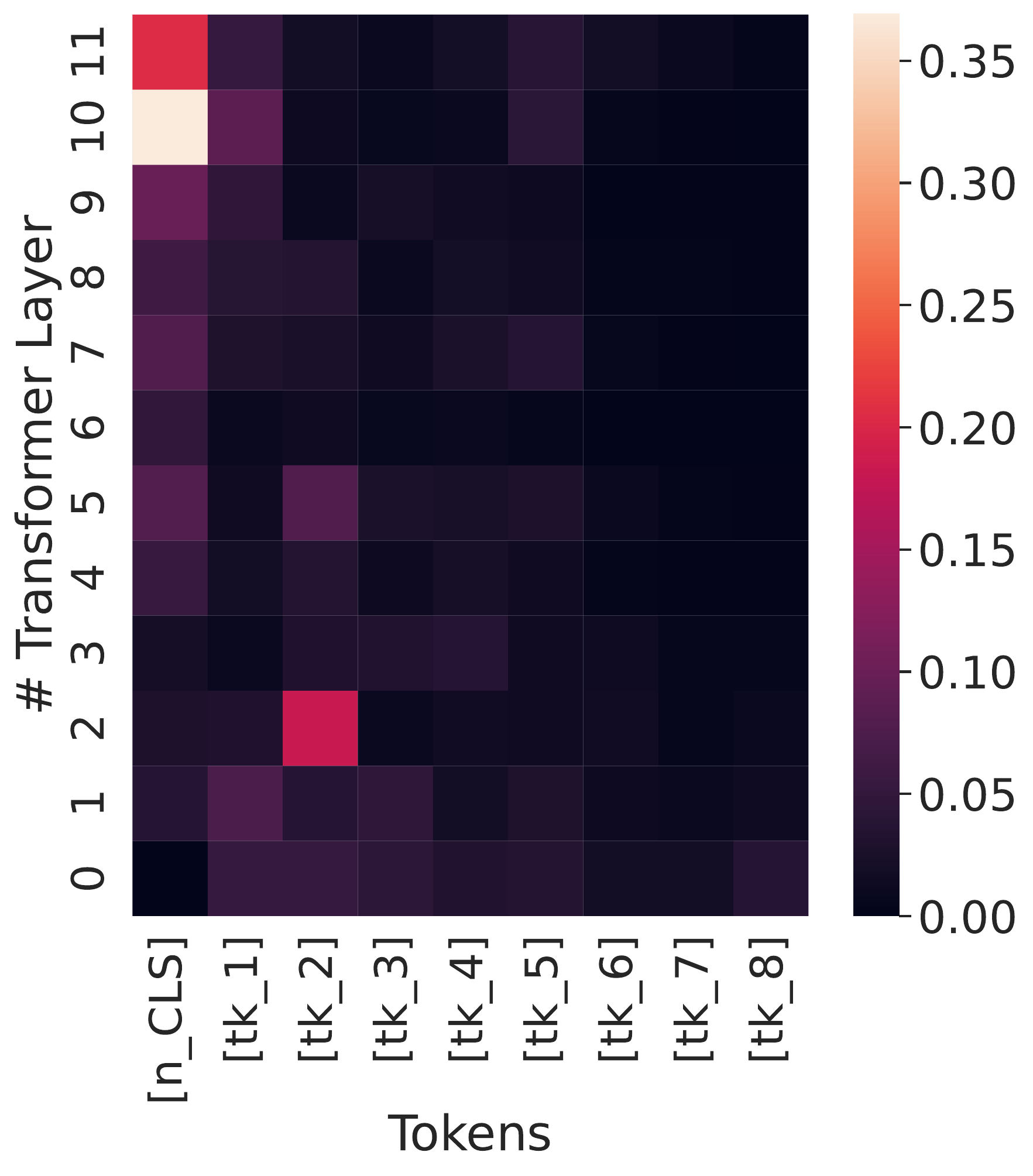}}
    \subfigure[Retrieval w/ pretrain]{\includegraphics[width=0.24\textwidth]{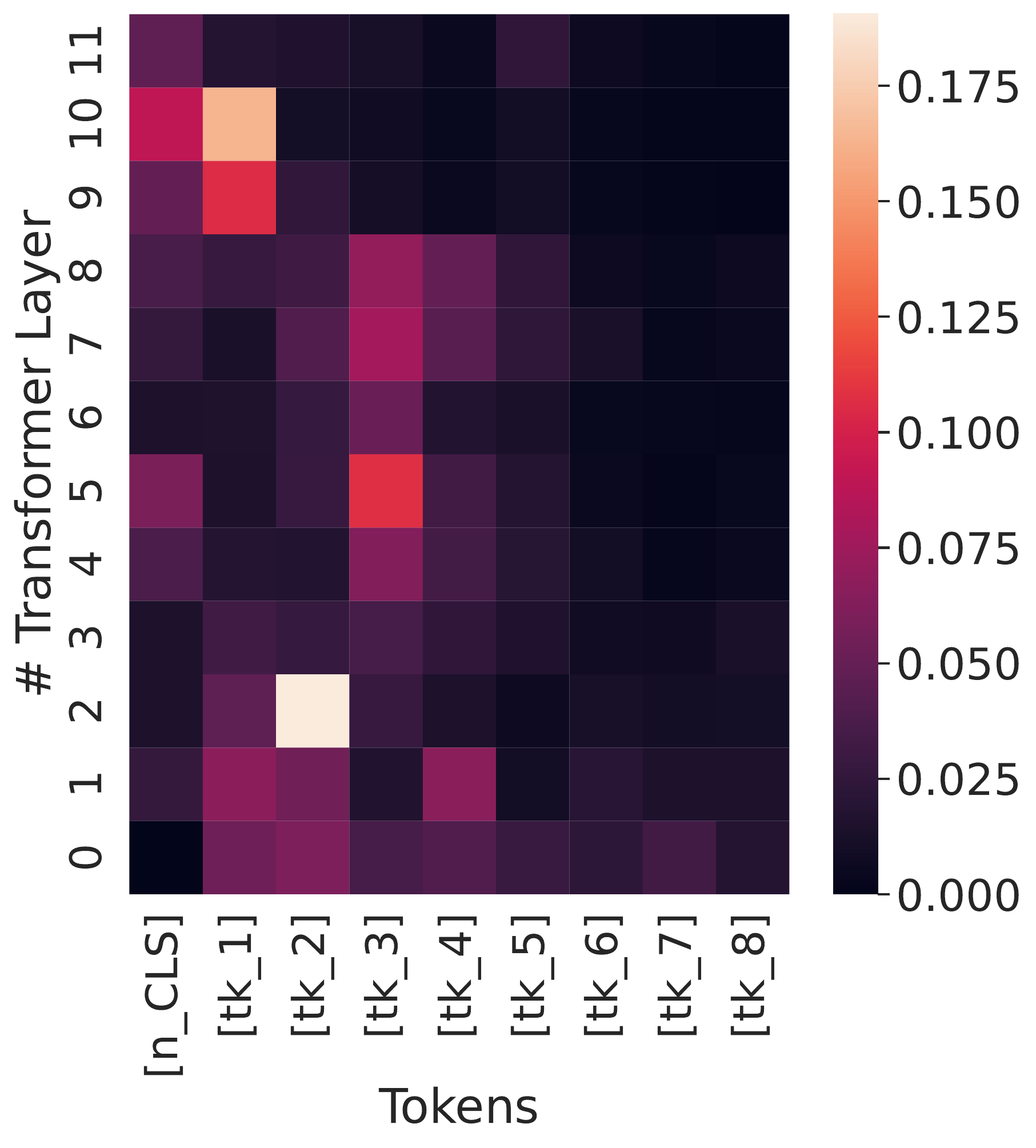}} 
    \subfigure[Reranking w/ pretrain]{\includegraphics[width=0.24\textwidth]{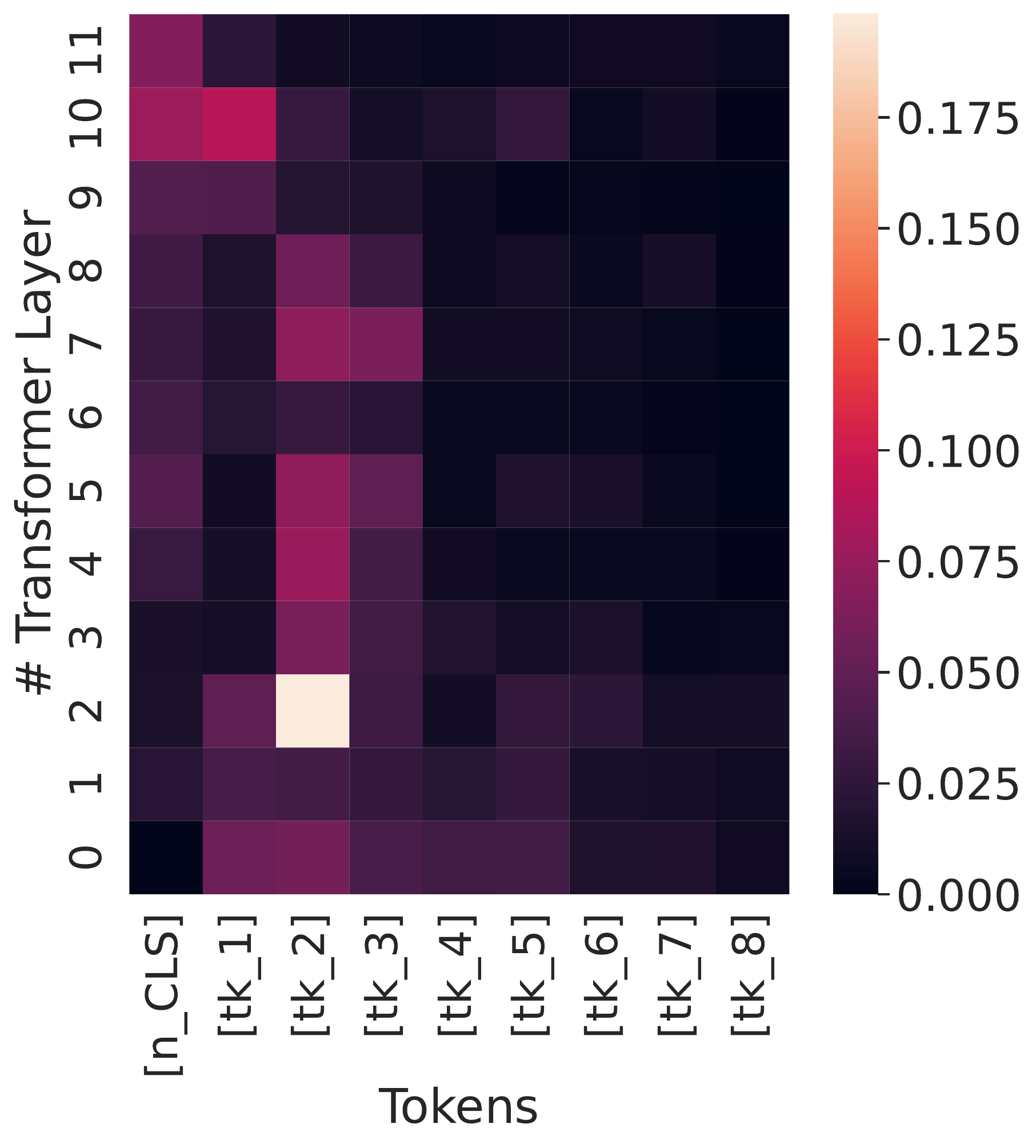}}
    \subfigure[Link Prediction w/ pretrain]{\includegraphics[width=0.24\textwidth]{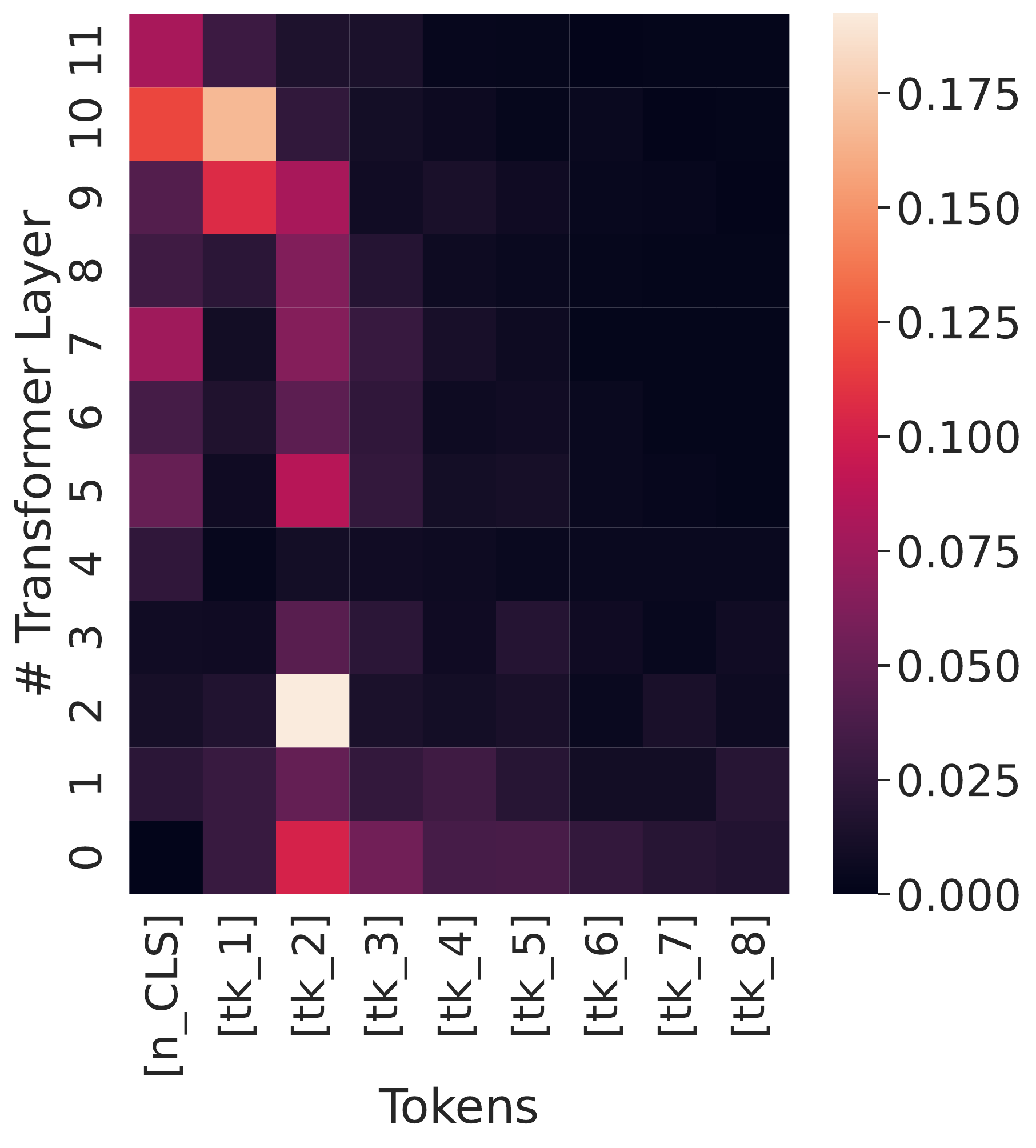}}
    \caption{Attention map study on Amazon-Sports. [n\_CLS] refers to network neighbor virtual token and [tk\_x]s refer to word tokens. [n\_CLS] is more activated after pretraining (\Ours), which means that more useful information from network neighbors is extracted to enhance center node text encoding.}
    \label{fig:attmap}
\end{figure*}

We conduct a case study by showing some attention maps of \Ours and the model without pretraining on four downstream tasks on Sports. 
We randomly pick a token from a random sample and plot the self-attention probability of how different tokens (x-axis), including neighbor virtual token ([n\_CLS]) and the first eight original text tokens ([tk\_x]), will contribute to the encoding of this random token in different layers (y-axis).
The result is shown in Figure \ref{fig:attmap}. 
From the result, we can find that the neighbor virtual token is more deactivated for the model without pretraining, which means that the information from neighbors is not fully utilized during encoding. However, the neighbor virtual token becomes more activated after pretraining, bringing more useful information from neighbors to enhance center node text encoding.

\section{Related Work}

\subsection{Pretrained Language Models}

Pretrained language models have been very successful in natural language processing since they were introduced \cite{peters2018deep,devlin2018bert}. 
Follow-up research has made them stronger by scaling them up from having millions of parameters \cite{yang2019xlnet, lewis2019bart, clark2020electra} to even trillions \cite{radford2019language, raffel2020exploring, brown2020language}. 
Another way that these models have been improved is by using different training objectives, including masked language modeling \cite{devlin2018bert}, auto-regressive causal language modeling \cite{brown2020language}, permutation language modeling \cite{yang2019xlnet}, discriminative language modeling \cite{clark2020electra}, correcting and contrasting \cite{meng2021coco} and document relation modeling \cite{yasunaga2022linkbert}. 
However, most of them are designed for modeling texts exclusively, and do not consider the inter-document structures.
In this paper, we innovatively design strategies to capture the semantic hints hidden inside the complex document networks.

\subsection{Domain Adaptation in NLP}

Large language models have demonstrated their power in various NLP tasks. However, their performance under domain shift is quite constrained \cite{ramponi2020neural}. To overcome the negative effect caused by domain shift, continuous pretraining is proposed in recent works \cite{gururangan2020don}, which can be further categorized into domain-adaptive pretraining \cite{han2019unsupervised} and task-specific pretraining \cite{howard2018universal}. 
However, existing works mainly focus on continuous pretraining based on textual information, while our work tries to conduct pretraining utilizing textual signal and network structure signal simultaneously.

\subsection{Pretraining on Graphs}
Inspired by the recent success of pretrained language models, researchers are starting to explore pretraining strategies for graph neural networks \cite{hu2020gpt, qiu2020gcc, hu2019strategies}. Famous strategies include graph autoregressive modeling \cite{hu2020gpt}, masked component modeling \cite{hu2019strategies}, graph context prediction \cite{hu2019strategies} and constrastive pretraining \cite{qiu2020gcc, velickovic2019deep, sun2019infograph}. 
These works conduct pretraining for graph neural network utilizing network structure information and do not consider the associated rich textual signal. 
However, our work proposes to pretrain the language model, adopting both textual information and network structure information.

\section{Conclusions}
In this work, we introduce \Ours, a method to pretrain language models on text-rich networks.
\Ours consists of two objectives: (1) a network-contextualized MLM pretraining objective and (2) a masked node prediction objective, to capture the rich semantics information hidden inside the complex network structure.
We conduct experiment on four downstream tasks and five datasets from two different domains, where \Ours outperforms baselines significantly and consistently.

\section*{Acknowledgments}
We thank anonymous reviewers for their valuable and insightful feedback. 
Research was supported in part by US DARPA KAIROS Program No. FA8750-19-2-1004 and INCAS Program No. HR001121C0165, National Science Foundation IIS-19-56151, IIS-17-41317, and IIS 17-04532, and the Molecule Maker Lab Institute: An AI Research Institutes program supported by NSF under Award No. 2019897, and the Institute for Geospatial Understanding through an Integrative Discovery Environment (I-GUIDE) by NSF under Award No. 2118329. Any opinions, findings, and conclusions or recommendations expressed herein are those of the authors and do not necessarily represent the views, either expressed or implied, of DARPA or the U.S. Government.
The views and conclusions contained in this paper are those of the authors and should not be interpreted as representing any funding agencies.

\section*{Limitations}

In this work, we mainly focus on language model pretraining on homogeneous text-rich networks and explore how pretraining can benefit classification, retrieval, reranking, and link prediction. Interesting future studies include 1) researching how to conduct pretraining on heterogeneous text-rich networks and how to characterize the edges of different semantics; 2) exploring how pretraining can benefit broader task spaces including summarization and question answering.

\section*{Ethics Statement}
While it has been shown that PLMs are powerful in language understanding \cite{devlin2018bert, lewis2019bart, raffel2020exploring}, there are studies highlighting their drawbacks such as the presence of social bias \cite{liang2021towards} and misinformation \cite{abid2021persistent}. In our work, we focus on pretraining PLMs with information from the inter-document structures, which could be a way to mitigate bias and eliminate the contained misinformation.

\bibliography{anthology,custom}
\bibliographystyle{acl_natbib}

\newpage

\appendix

\section{Pretrain Settings}\label{App:pretrain_setup}
To facilitate the reproduction of our pretraining experiment, we provide the hyperparameter configuration in Table \ref{apx:tab:pretrain}.
All reported continuous pretraining and in-domain pretraining methods use exactly the same set of hyperparameters for pretraining for a fair comparison.
All GraphFormers \cite{yang2021graphformers} involved methods have the neighbor sampling number set as 5.
Paper titles and item titles are used as text associated with the nodes in the two kinds of networks, respectively. (For some items, we concatenate the item title and description together since the title is too short.) Since most paper titles (88\%) and item titles (97\%) are within 32 tokens, we set the max length of the input sequence to be 32.
The models are trained for 5/10/30 epochs (depending on the size of the network) on 4 Nvidia A6000 GPUs with a total batch size of 512. The total time cost is around 24 hours for each network.
Code is available at \url{https://github.com/PeterGriffinJin/Patton}.

\begin{table*}[ht]
\centering
\caption{Pretraining hyper-parameter configuration.}\label{apx:tab:pretrain}
\scalebox{0.78}{
\begin{tabular}{cccccc}
\toprule
Parameter & Mathematics & Geology & Economics & Clothes & Sports \\
\midrule
Max Epochs & 30 & 10 & 30 & 5 & 10 \\
Peak Learning Rate & 1e-5 & 1e-5 & 1e-5 & 1e-5 & 1e-5 \\
Batch Size & 512 & 512 & 512 & 512 & 512 \\
Warm-Up Epochs & 3 & 1 & 3 & 0.5 & 1 \\
Sequence Length & 32 & 32 & 32 & 32 & 32 \\
Adam $\epsilon$ & 1e-8 & 1e-8 & 1e-8 & 1e-8 & 1e-8 \\
Adam $(\beta_1, \beta_2)$ & (0.9, 0.999) & (0.9, 0.999) & (0.9, 0.999) & (0.9, 0.999) & (0.9, 0.999) \\
Clip Norm & 1.0 & 1.0 & 1.0 & 1.0 & 1.0 \\
Dropout & 0.1 & 0.1 & 0.1 & 0.1 & 0.1 \\
\bottomrule            
\end{tabular}
}
\end{table*}

\section{Classification}\label{apx:class}

\paragraph{Task.}
The coarse-grained category names for academic networks and e-commerce networks are the first-level category names in the network-associated category taxonomy. 
We train all the methods in the 8-shot setting (8 labeled training samples and 8 labeled validation samples for each class) and test the models with hundreds of thousands of new query nodes (220,681, 215,148, 85,346, 477,700, and 129,669 for Mathematics, Geology, Economics, Clothes, and Sports respectively).
Detailed information on all category names can be found in Table \ref{tab:math}-\ref{tab:sports}.

\paragraph{Finetuning Settings.}
All reported methods use exactly the same set of hyperparameters for finetuning for a fair comparison. The median results of three runs with the same set of three different random seeds are reported. 
For all the methods, we finetune the model for 500 epochs in total. The peak learning rate is 1e-5, with the first $10\%$ steps as warm-up steps. The training batch size and the validation batch size are both 256. During training, we validate the model every 25 steps and the best checkpoint is utilized to perform prediction on the test set.
The experiments are carried out on one Nvidia A6000 GPU.

\begin{table*}[t]
\caption{Class names of MAG-Mathematics.}\label{tab:math}
\resizebox{\textwidth}{!}{
\begin{tabular}{llllllll}
\toprule
0 & mathematical optimization & 5 & econometrics & 10 & control theory & 15 & computational science \\
1 & mathematical analysis & 6 & mathematical physics & 11 & geometry & 16 & mathematics education \\
2 & combinatorics & 7 & statistics & 12 & applied mathematics & 17 & arithmetic \\
3 & algorithm & 8 & pure mathematics & 13 & operations research & &  \\
4 & algebra & 9 & discrete mathematics & 14 & mathematical economics &  &  \\
\bottomrule
\end{tabular}}
\end{table*}

\begin{table*}[t]
\caption{Class names of MAG-Geology.}\label{tab:geology}
\resizebox{\textwidth}{!}{
\begin{tabular}{llllllll}
\toprule
0 & geomorphology & 5 & paleontology & 10 & petrology & 15 & mining engineering \\
1 & seismology & 6 & climatology & 11 & geotechnical engineering & 16 & petroleum engineering \\
2 & geochemistry & 7 & atmospheric sciences & 12 & soil science &  &  \\
3 & mineralogy & 8 & geodesy & 13 & earth science &  &  \\
4 & geophysics & 9 & oceanography & 14 & remote sensing &  &  \\
\bottomrule
\end{tabular}}
\end{table*}

\begin{table*}[t]
\caption{Class names of the MAG-Economics}\label{tab:economy}
\resizebox{\textwidth}{!}{
\begin{tabular}{llllllll}
\toprule
0 & mathematical economics & 10 & economy                 & 20 & development economics      & 30 & economic policy                   \\
1 & labour economics       & 11 & monetary economics      & 21 & international trade        & 31 & market economy                    \\
2 & finance                & 12 & operations management   & 22 & keynesian economics        & 32 & environmental economics           \\
3 & econometrics           & 13 & actuarial science       & 23 & positive economics         & 33 & classical economics               \\
4 & macroeconomics         & 14 & industrial organization & 24 & agricultural economics     & 34 & management science                \\
5 & microeconomics         & 15 & political economy       & 25 & international economics    & 35 & management                        \\
6 & economic growth        & 16 & commerce                & 26 & demographic economics      & 36 & welfare economics                 \\
7 & financial economics    & 17 & socioeconomics          & 27 & neoclassical economics     & 37 & economic system                   \\
8 & public economics       & 18 & financial system        & 28 & natural resource economics & 38 & environmental resource management \\
9 & law and economics      & 19 & accounting              & 29 & economic geography         & 39 & economic history   \\
\bottomrule
\end{tabular}}
\end{table*}

\begin{table*}[t]
\caption{Class names of Amazon-Clothes.}\label{tab:cloth}
\begin{tabular}{lp{0.2\textwidth}lp{0.2\textwidth}lp{0.2\textwidth}lp{0.2\textwidth}}
\toprule
0 & girls & 3 & luggage & 5 & fashion watches & 7 & boys   \\
1 & men & 4 & baby & 6 & shoes & 8 & adidas \\
2 & novelty &   &         &   &                 &   &       \\
\bottomrule
\end{tabular}
\end{table*}

\begin{table*}[t]
\caption{Class name of Amazon-Sports.}\label{tab:sports}
\resizebox{\textwidth}{!}{
\begin{tabular}{llllllll}
\toprule
0 & accessories             & 4 & cycling                    & 8  & golf                            & 12 & paintball \& airsoft \\
1 & action sports           & 5 & baby                       & 9  & hunting \& fishing \& game room & 13 & racquet sports       \\
2 & boating \& water sports & 6 & exercise \& leisure sports & 10 & outdoor gear                    & 14 & snow sports          \\
3 & clothing                & 7 & fan shop                   & 11 & fitness                         & 15 & team sports      \\
\bottomrule
\end{tabular}}
\end{table*}

\section{Retrieval}\label{apx:retrieval}
\paragraph{Task.} The retrieval task corresponds to fine-grained category retrieval. Given a node in the network, we aim to retrieve its fine-grained labels from a large label space. 
We train all the compared methods in the 16-shot setting (16 labeled queries in total) and test the models with tens of thousands of new query nodes (38,006, 33,440, 14,577, 95,731, and 34,979 for Mathematics, Geology, Economics, Clothes, and Sports, respectively).
The fine-grained label spaces for both academic networks and e-commerce networks are constructed from all the labels in the network-associated taxonomy \footnote{https://www.microsoft.com/en-us/research/project/academic/articles/visualizing-the-topic-hierarchy-on-microsoft-academic/} \footnote{http://jmcauley.ucsd.edu/data/amazon/links.html}. The statistics of the label space for all networks can be found in Table \ref{tab:dataset}.

\paragraph{Finetuning Settings.}
We finetune the models with the widely-used DPR pipeline \cite{karpukhin2020dense}. All reported methods use exactly the same set of hyperparameters for finetuning for a fair comparison. The median results of three runs with the same set of three different random seeds are reported. 
For all the methods, we finetune the model for 1,000 epochs with the training data. The peak learning rate is 1e-5, with the first $10\%$ steps as warm-up steps. The training batch size is 128. The number of hard BM25  negative samples\footnote{https://github.com/dorianbrown/rank\_bm25} is set as 4. We utilize the faiss library \footnote{https://github.com/facebookresearch/faiss} to perform an approximate search for nearest neighbors. 
The experiments are carried out on one Nvidia A6000 GPU.

\section{Reranking}\label{apx:reranking}
\paragraph{Task.} The reranking task corresponds to fine-grained category reranking. Given a retrieved category list for the query node, we aim to rerank all categories within the list. 
We train all the methods in the 32-shot setting (32 training queries and 32 validation queries) and test the models with 10,000 new query nodes and candidate list pairs.
The category space in reranking is the same as that in retrieval. In our experiment, the retrieved category list is constructed with BM25 and exact matching of category names.

\paragraph{Finetuning Settings.}
All reported methods use exactly the same set of hyperparameters for finetuning for a fair comparison. The median results of three runs with the same set of three different random seeds are reported.
For all the methods, we finetune the model for 1,000 epochs in total with the training data. The peak learning rate is 1e-5, with the first $10\%$ steps as warm-up steps. The training batch size and validation batch size are 128 and 256, respectively. During training, the model is validated every 1,000 steps and the best checkpoint is utilized to conduct inference on the test set.
The experiments are carried out on one Nvidia A6000 GPU.

\section{Link Prediction}\label{apx:lp}
\paragraph{Task.}
The task aims to predict if there should exist an edge with specific semantics between two nodes. It is worth noting that the semantics of the edge here is different from the semantics of the edge in the pretraining text-rich network. In academic networks, the edge semantics in the pretraining network is ``citation'', while the edge semantics in downstream link prediction is ``author overlap'' \footnote{Two papers have at least one same author.}. In e-commerce networks, the edge semantics in the pretraining network is ``co-viewed'', while the edge semantics in the prediction of the downstream link is ``co-purchased''. We train all the methods in the 32-shot setting (32 training labeled pairs and 32 validation labeled pairs) and test the models with 10,000 new node pairs. We utilize in-batch samples as negative samples in training to finetune the model and in testing to evaluate the performance of the methods.

\paragraph{Finetuning Settings.}
All reported methods use exactly the same set of hyperparameters for finetuning for a fair comparison. The median results of three runs with the same set of three different random seeds are reported.
For all the methods, we finetune the model for 200 epochs in total. The peak learning rate is 1e-5, with the first $10\%$ step as warm-up steps. The training batch size and validation batch size are 128 and 256, respectively. During training, we validate the model in 20 steps and use the best checkpoint to perform the prediction on the test set.
The experiments are carried out on one Nvidia A6000 GPU.

\section{Finetuning Data Size Study}\label{apx:finetune}
\begin{figure*}[t]
    \centering
    \subfigure[Classification]{\includegraphics[width=0.24\textwidth]{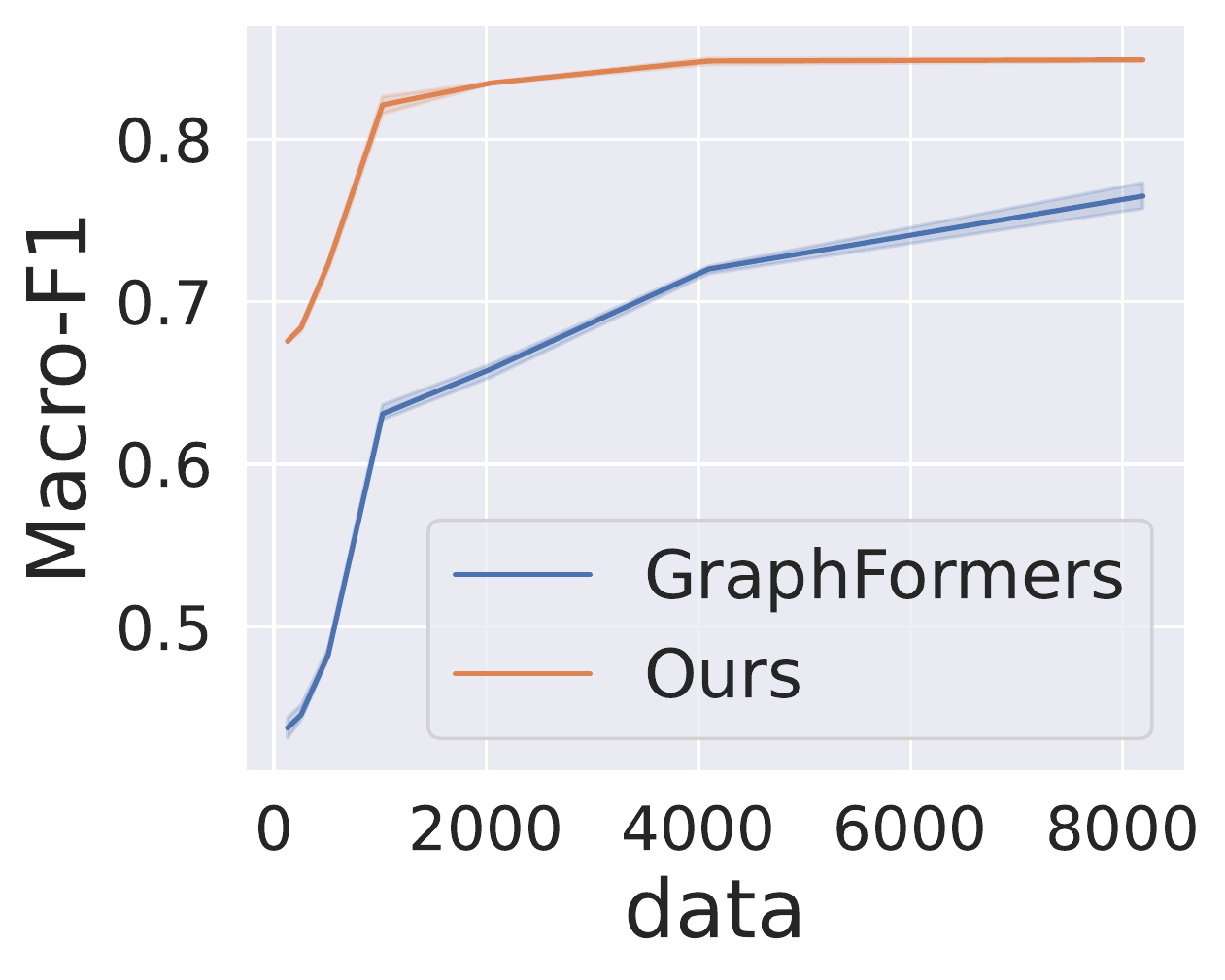}}
    \subfigure[Retrieval]{\includegraphics[width=0.24\textwidth]{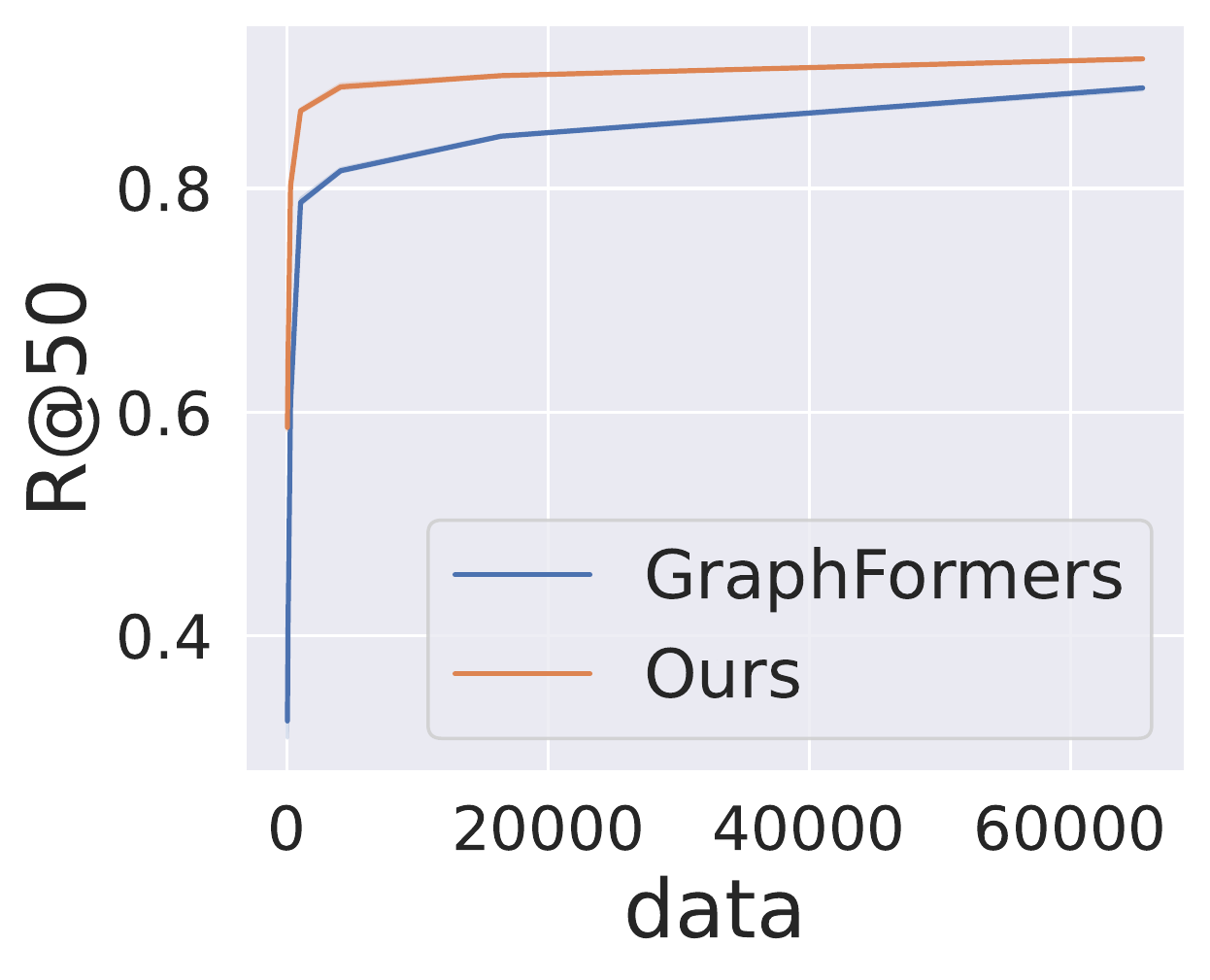}} 
    \subfigure[Reranking]{\includegraphics[width=0.24\textwidth]{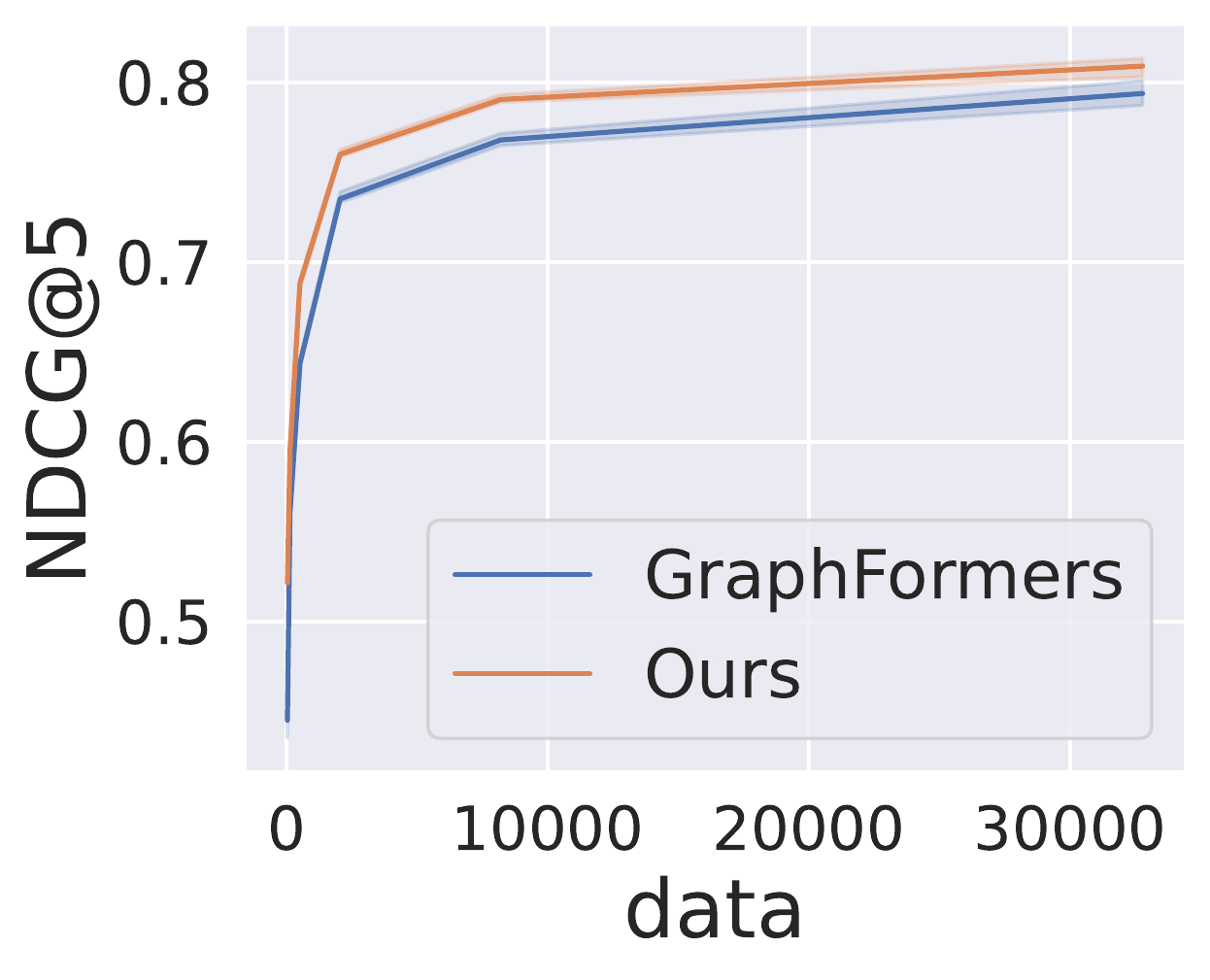}}
    \subfigure[Link Prediction]{\includegraphics[width=0.24\textwidth]{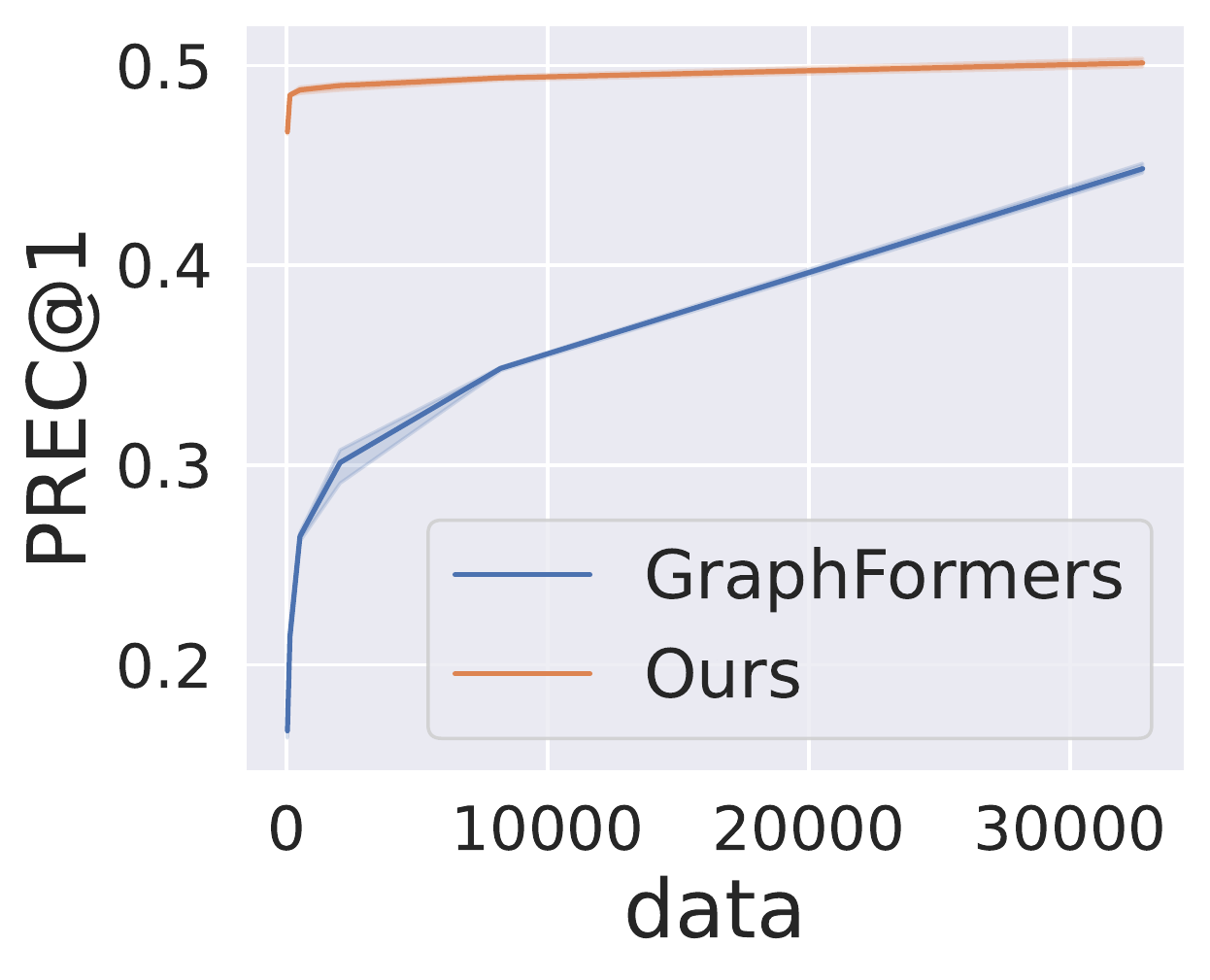}}
    \caption{Finetuning data size study on Amazon-Sports. As finetuning data size increases, both the performance of our proposed \Ours and the model without pretraining (GraphFormers) improves. \Ours consistently outperforms the language model without pretraining (GraphFormers).}
    \label{fig:finetune-data}
\end{figure*}

We conduct a parameter study to explore how beneficial our pretraining method is to downstream tasks with different amounts of finetuning data on the four tasks on Sports.
The results are shown in Figure \ref{fig:finetune-data}, where we can find that: 1) As finetuning data increases, the performance of both \Ours and the model without pretraining (GraphFormers) improves. 2) The performance gap between \Ours and the model without pretraining (GraphFormers) becomes smaller as finetuning data increases, but \Ours is consistently better than the model without pretraining (GraphFormers).

\end{document}